%% file: AAMAS_2026_main.tex
\documentclass[sigconf]{aamas}

\usepackage{balance} 
\usepackage{multirow}
\usepackage{adjustbox}
\usepackage{pifont}
\newcommand{\xmark}{\ding{55}}
\newcommand\ph{$\phantom{1}$}

\usepackage{algorithm}
\usepackage{algpseudocode}
\usepackage{tabularx}
\usepackage{listings}
\definecolor{codegray}{rgb}{0.95,0.95,0.95}
\definecolor{keywordblue}{rgb}{0.13,0.13,1}
\definecolor{commentgreen}{rgb}{0,0.5,0}
\definecolor{stringpurple}{rgb}{0.58,0,0.82}
\doi{UCGT7089}

\makeatletter
\gdef\@copyrightpermission{
  \begin{minipage}{0.2\columnwidth}
   \href{https://creativecommons.org/licenses/by/4.0/}{\includegraphics[width=0.90\textwidth]{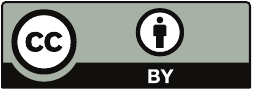}}
  \end{minipage}\hfill
  \begin{minipage}{0.8\columnwidth}
   \href{https://creativecommons.org/licenses/by/4.0/}{This work is licensed under a Creative Commons Attribution International 4.0 License.}
  \end{minipage}
  \vspace{5pt}
}
\makeatother

\setcopyright{ifaamas}
\acmConference[AAMAS '26]{Proc.\@ of the 25th International Conference
on Autonomous Agents and Multiagent Systems (AAMAS 2026)}{May 25 -- 29, 2026}
{Paphos, Cyprus}{C.~Amato, L.~Dennis, V.~Mascardi, J.~Thangarajah (eds.)}
\copyrightyear{2026}
\acmYear{2026}
\acmDOI{}
\acmPrice{}
\acmISBN{}

\title[AAMAS-2026 Formatting Instructions]{ReAcTree: Hierarchical LLM Agent Trees with Control Flow for Long-Horizon Task Planning}

\author{Jae-Woo Choi}
\affiliation{
  \institution{ETRI \& UST}
  \city{Daejeon}
  \country{Republic of Korea}}
\email{jwchoi0717@etri.re.kr}

\author{Hyungmin Kim}
\affiliation{
  \institution{UST}
  \city{Daejeon}
  \country{Republic of Korea}}
\email{khm159@etri.re.kr}

\author{Hyobin Ong}
\affiliation{
  \institution{UST}
  \city{Daejeon}
  \country{Republic of Korea}}
\email{ohnghb@etri.re.kr}

\author{Youngwoo Yoon}
\affiliation{
  \institution{ETRI \& UST}
  \city{Daejeon}
  \country{Republic of Korea}}
\email{youngwoo@etri.re.kr}

\author{Minsu Jang}
\affiliation{
  \institution{ETRI \& UST}
  \city{Daejeon}
  \country{Republic of Korea}}
\email{minsu@etri.re.kr}

\author{Dohyung Kim}
\affiliation{
  \institution{ETRI \& UST}
  \city{Daejeon}
  \country{Republic of Korea}}
\email{dhkim008@etri.re.kr}

\author{Jaehong Kim}
\affiliation{
  \institution{ETRI}
  \city{Daejeon}
  \country{Republic of Korea}}
\email{jhkim504@etri.re.kr}

\begin{abstract}
Recent advancements in large language models (LLMs) have enabled significant progress in decision-making and task planning for embodied autonomous agents. However, most existing methods struggle with complex, long-horizon tasks because they rely on a monolithic trajectory that entangles all past decisions and observations to solve the entire task in a single unified process. To address this limitation, we propose \emph{ReAcTree}, a hierarchical task-planning method that decomposes a complex goal into manageable subgoals within a dynamically constructed agent tree. Each subgoal is handled by an LLM agent node capable of reasoning, acting, and further expanding the tree, while control flow nodes coordinate the execution strategies of agent nodes. In addition, we integrate two complementary memory systems: each agent node retrieves goal-specific, subgoal-level examples from \emph{episodic memory} and shares environment-specific observations through \emph{working memory}. Experiments on the WAH-NL and ALFRED show \emph{ReAcTree} consistently outperforms strong task-planning baselines such as \emph{ReAct} across diverse LLMs. Notably, on WAH-NL, \emph{ReAcTree} achieves a 61\% goal success rate with Qwen 2.5 72B, nearly doubling \emph{ReAct}’s 31\%. The code is available at \href{https://github.com/Choi-JaeWoo/ReAcTree.git}{https://github.com/Choi-JaeWoo/ReAcTree.git}.
\end{abstract}

\keywords{Hierarchical Task Planning; LLM Agents; Behavior Trees}

\newcommand{\BibTeX}{\rm B\kern-.05em{\sc i\kern-.025em b}\kern-.08em\TeX}

\begin{document}


\pagestyle{fancy}
\fancyhead{}


\maketitle 

\input{latex/01_introduction}
\input{latex/02_related_work}
\input{latex/03_Preliminaries}
\input{latex/04_reactree}
\input{latex/05_experiments}

\input{latex/06_conclusion}




\begin{acks}
This work was supported by the Institute of Information \& Communications Technology Planning \& Evaluation (IITP) grant funded by the Korea government (MSIT) (RS-2024-00336738, Development of Complex Task Planning Technologies for Autonomous Agents, 40\% and RS-2022-II220951, Development of Uncertainty-Aware Agents Learning by Asking Questions, 40\%), and the National Research Council of Science \& Technology (NST) grant by the Korea government (MSIT) (No. GTL25041-000, 20\%).
\end{acks}

\balance

\bibliographystyle{ACM-Reference-Format} 
\bibliography{reference}


\clearpage
\onecolumn
\appendix
\input{appendix}

\end{document}

%% file: latex/01_introduction.tex
\section{Introduction}
\label{introduction}

Large language models (LLMs) have recently exhibited remarkable reasoning capabilities by generating intermediate reasoning steps \cite{wei2022chain, kojima2022large}. These advances open up new possibilities for decision-making and task planning in embodied autonomous agents, with the long-term objective of enabling them to fulfill high-level natural language commands autonomously. Early works \cite{huang2022language, ahn2022can, song2023llm, liang2023code} have demonstrated that LLMs can leverage their pre-trained world knowledge to generate mid-level action sequences without additional training, primarily through prompting rather than handcrafted heuristics or learned policies \cite{eysenbach2019search, xu2019regression}.

The decision-making and planning capabilities of LLMs can be further improved by incorporating observations or feedback from external environments. Recent approaches explore adapting the next action \cite{yao2023react, huang2022inner, zeng2023socratic}, adjusting the entire plan \cite{shinn2024reflexion}, or refining both the plan and the next action \cite{wang2023describe, sun2023adaplanner}. However, these methods often use a monolithic trajectory that entangles all past decisions and observations from multiple subgoals, increasing the risk of hallucination and logical failures \cite{liu2024lost}. Another line of research investigates multiple reasoning paths \cite{wang2023selfconsistency, yao2024tree, zhou2024language} for general reasoning tasks. Such methods typically assume the capability to simulate multiple paths and revert to previous states \cite{zhao2024large, zhuang2024toolchain, hu2024treeplanner}, which is rarely feasible in real-world scenarios due to irreversible actions (e.g., slicing an apple) and partial observability.

\begin{figure*}[t!]
    \centering
    \includegraphics[width=\textwidth]{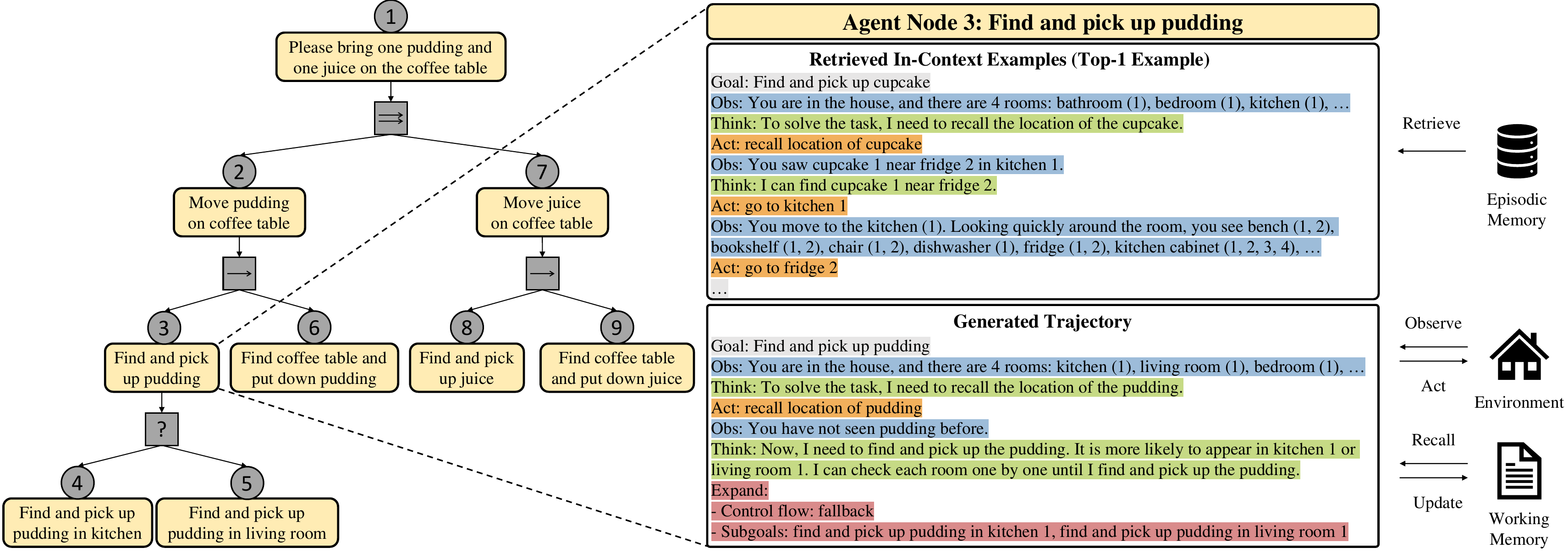}
    \caption{Illustration of how \emph{ReAcTree} generates an agent tree for the instruction: \textit{Please bring one pudding and one juice to the coffee table.} The left side shows the hierarchical structure with agent nodes (circles) and control flow nodes (squares); the number inside each circle denotes execution order, and the attached text box specifies the corresponding subgoal. Control flow nodes are labeled by type ($\rightarrow$ for sequence, ? for fallback, and $\Rightarrow$ for parallel). The right side highlights the decision-making trajectory of agent node 3, including observation, reasoning, acting, expansion, and episodic and working memory usage.}
    \label{fig:fig1}
    \Description{The figure is divided into two parts illustrating the ReAcTree framework. The left side shows the hierarchical decomposition of a complex task into an agent tree using control flow nodes like sequence, fallback, and parallel. The right side zooms in on a single agent node to demonstrate its internal decision-making process, which involves retrieving examples from episodic memory, reasoning to generate actions, and dynamically expanding the tree structure.}
\end{figure*}

To tackle these challenges, we introduce \emph{ReAcTree}, a novel framework that dynamically constructs an agent tree in the subgoal space rather than an action tree over primitive actions. Each LLM-powered agent node is responsible for a subgoal and extends the \emph{ReAct} paradigm \cite{yao2023react}: it can reason, act, or expand the tree by proposing new subgoals with an associated control flow when a task is too complex. Control flow nodes, inspired by Behavior Trees \cite{colledanchise2018learning}, coordinate these agents by sequencing, fallback, or parallel execution. Conceptually, \emph{ReAcTree} extends the \emph{Least-to-Most} prompting strategy \cite{zhou2023least} from static reasoning to dynamic, agentic planning. By decomposing a problem into semantically isolated subgoals and providing each with its own focused context, \emph{ReAcTree} prevents the propagation of reasoning errors and makes long-horizon tasks more tractable for LLMs. The hierarchical structure of \emph{ReAcTree} is illustrated in Figure~\ref{fig:fig1}.

To enhance in-context learning and coordination capabilities, we incorporate two complementary memory systems. First, \emph{episodic memory} assists each agent node in effective in-context learning by providing subgoal-level examples from past experiences that are semantically similar to the current subgoal. It can be bootstrapped from a small set of manually collected trajectories and gradually expanded through successful runs. Second, \emph{working memory} functions as a shared blackboard, allowing agent nodes to exchange critical observations (e.g., the locations of movable objects). This collective situational awareness reduces redundant searches and mitigates hallucinations. Together, these memory systems enable \emph{ReAcTree} to learn from the past and coordinate in the present, strengthening the robustness of agentic decision-making under partial observability.

We evaluate \emph{ReAcTree} by extending LoTa-Bench \cite{choi2024lotabench}, a benchmark providing two household task environments (WAH-NL \cite{puig2021watchandhelp, choi2024lotabench} with VirtualHome \cite{puig2018virtualhome} and ALFRED \cite{shridhar2020alfred} with AI2THOR \cite{kolve2017ai2}), to a more challenging partially observable setting. This setup reflects real-world constraints and tests agentic decision-making under uncertainty. Across both environments, \emph{ReAcTree} consistently outperforms strong baselines across various LLMs. On WAH-NL, it achieves a 61\% goal success rate with Qwen 2.5 72B, nearly doubling \emph{ReAct}’s 31\% under the same model size. Notably, even with a much smaller 7B model, it reaches 37\%, showing that \emph{ReAcTree} maintains strong performance despite reduced model capacity. The contributions of its core components are further substantiated by ablation studies on memory systems and control flow types, as well as in-depth analyses of computational cost and failure modes.

In summary, this paper makes the following contributions: (1) \emph{ReAcTree}, a hierarchical planning framework for long-horizon tasks that dynamically constructs an LLM agent tree, where each agent node independently solves a subgoal while control flow nodes coordinate their execution into a coherent plan; (2) two memory systems: episodic memory that enhances in-context learning ability of agent nodes, and working memory that facilitates information sharing across agent nodes to support collective situational awareness; and (3) extensive experiments on LoTa-Bench under partially observable settings, demonstrating the effectiveness of \emph{ReAcTree} across strong baselines and providing detailed analysis through ablations, computational cost analysis, and failure case studies.

%% file: latex/02_related_work.tex
\begin{figure*}[t!]
    \centering
    \includegraphics[width=0.8\textwidth]{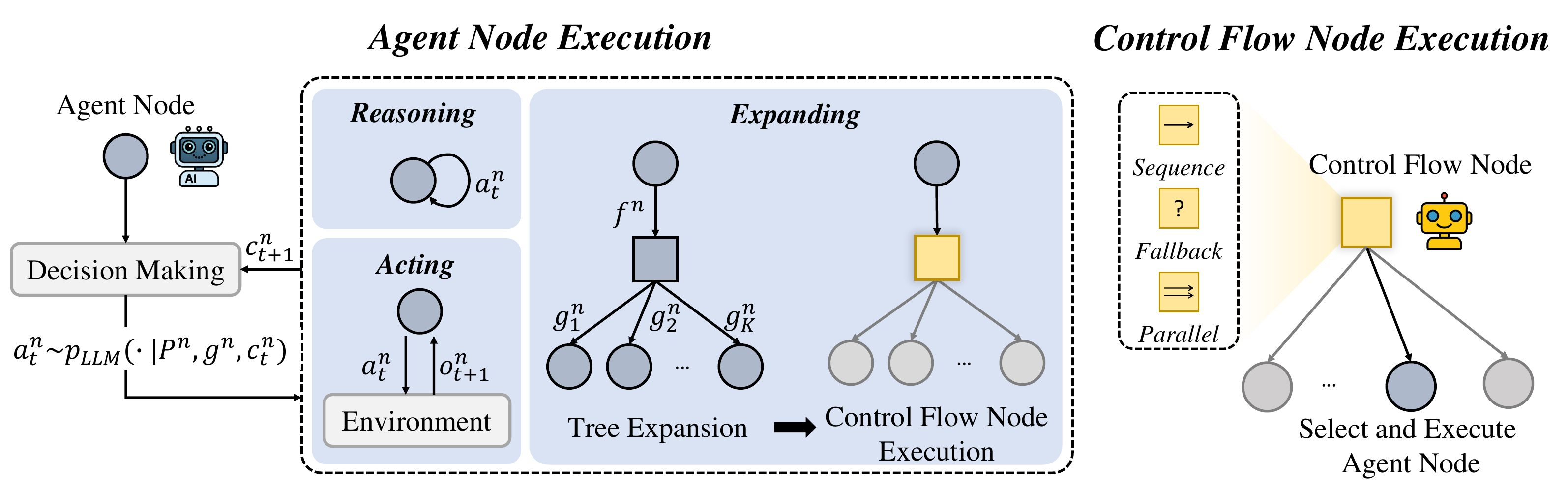}
    \caption{Illustration of agent node execution and control flow node execution in \emph{ReAcTree}.}
    \label{fig:reactree_architecture}
    \Description{The figure diagrams the execution cycle in ReAcTree. It shows how an Agent Node chooses between reasoning, acting, or expanding into a subtree, and how the resulting Control Flow Node (Sequence, Fallback, or Parallel) manages the execution of child nodes.}
\end{figure*}

\section{Related Work}
\label{relatedworks}
\textbf{LLM-based Embodied Agents.}
LLMs such as GPT-3, GPT-4, PaLM, LLaMA, and Qwen \cite{brown2020language, achiam2023gpt, chowdhery2023palm, dubey2024llama, yang2024qwen2} have demonstrated impressive few-shot in-context learning capabilities, advancing the state of the art in NLP tasks. Prompting techniques \cite{wei2022chain, kojima2022large, zhou2023least, gao2023pal} further enhance reasoning by encouraging articulation of intermediate steps. Building on these advancements, research has increasingly explored text-based applications to embodied decision-making. Early studies \cite{huang2022language, ahn2022can, song2023llm} showed that LLMs could generate mid-level action sequences without additional training, while later studies introduced code-based plan generation \cite{liang2023code, singh2023progprompt}, the extension of classical agent architectures \cite{ichida2024bdi}, and the integration of environmental feedback or tools \cite{huang2022inner, zeng2023socratic, shen2024hugginggpt, lu2024chameleon, chen2023open}. \emph{ReAct} \cite{yao2023react} enhanced planning by prompting explicit intermediate reasoning, while \emph{Reflexion} \cite{shinn2024reflexion} applied iterative self-refinement \cite{madaan2024self, welleck2023generating, paul2024refiner, chen2024teaching}, offering additional flexibility for long-horizon tasks.

\textbf{Hierarchical Task Planning with LLMs.}
To tackle more complex, long-horizon tasks, researchers have introduced hierarchical frameworks that decompose goals into manageable planning levels. \emph{AdaPlanner} \cite{sun2023adaplanner} and \emph{DEPS} \cite{wang2023describe} adopt bi-level hierarchies, refining both an overall plan and next-step decisions with environmental feedback. Other methods combine classical task-and-motion planning with learned low-level controllers guided by LLM reasoning \cite{wong2024learning, liu2024learning, ding2023task, shukla2024lgts, wang2025inclet}. Some works leverage structured formalisms, such as behavior trees \cite{wang2024mosaic, chen2024integrating, ao2025llm} or hierarchical task networks \cite{munoz2025chathtn}, to organize actions.
However, they typically rely on predefined structures or domain-specific routines. In contrast, our approach enables dynamic task decomposition, expanding subgoals in complex environments without being tied to a specific domain.

\textbf{Tree Search-Based Planning with LLMs.}
A prominent line of research explores multiple reasoning paths to evaluate diverse hypotheses before committing to a final solution. Such approaches \cite{wang2023selfconsistency, yao2024tree, zhou2024language, xie2024self, besta2024graph, hao2023reasoning} show that systematically branching reasoning steps significantly improves performance in reasoning tasks. Several studies extend this idea to agentic planning. For instance, \emph{LLM-MCTS} \cite{zhao2024large} and \emph{ToolChain*} \cite{zhuang2024toolchain} search action trees using Monte Carlo Tree Search and A* search, respectively. \emph{Tree-Planner} \cite{hu2024treeplanner} independently samples planning paths and merges them into an action tree, then makes decisions based on grounded observations at each node. However, these methods typically assume reversible actions (i.e., rolling back to prior states) in simulators, which limits their applicability in real-world settings. In contrast, \emph{ReAcTree} constructs a dynamically expanding LLM agent tree that decomposes complex goals into subgoals, delegates them to agent nodes, and coordinates their execution via behavior tree-inspired control flow, replacing search-based exploration with agent coordination.

%% file: latex/03_Preliminaries.tex
\section{Preliminaries}
\label{preliminaries}
\textbf{Problem Formulation.} We consider the agentic planning problem as a sequential decision-making problem aimed at achieving a goal $g$ expressed in natural language. At each time step $t$, the agent has access to the context $c_t=(o_1, a_1, o_2, a_2, \cdots, a_{t-1},o_t)$, where $o_i$ and $a_i$ represent the observation and action at time step $i$, respectively. The objective of the agent is to generate the next action $a_t$ based on the context $c_t$, with the aim of eventually achieving the goal $g$.

\textbf{ReAct \citep{yao2023react}.} \emph{ReAct} addresses this problem by interleaving reasoning and action execution using a pre-trained LLM $p_{LLM}$. The action policy is defined as $a_t \sim p_{LLM}(\cdot | P, g, c_t)$, where $P = (P_{sys}, P_{ic})$ is the initial prompt, composed of a system prompt $P_{sys}$ and in-context examples $P_{ic}$. The key idea is to use the augmented action space, $\hat{\mathcal{A}}_t = \mathcal{A}_t \cup \mathcal{L}$, where $\mathcal{A}_t$ is the set of executable skills available at time $t$, and $\mathcal{L}$ is the language space representing reasoning steps or thoughts. If $a_t \in \mathcal{A}_t$, the agent executes the action and obtains a text observation from the environment. If $a_t \in \mathcal{L}$, it is called a thought or reasoning trace, which aids in the logical inference of the LLM. In this case, the agent does not receive a new observation from the environment, i.e., $o_{t+1} = \phi$.

%% file: latex/04_reactree.tex
\section{ReAcTree}
\label{sec:reactree}
\emph{ReAcTree} is a hierarchical task-planning framework for agentic decision-making in complex environments. It dynamically constructs a tree of \emph{agent nodes} and \emph{control flow nodes}, as illustrated in Figure~\ref{fig:reactree_architecture}. Agent nodes operate as LLM-based task planners that can \emph{reason}, \emph{act}, and \emph{expand} the tree with new subgoals and appropriate control flow. Control flow nodes, inspired by behavior trees \citep{colledanchise2018learning}, determine how child agent nodes are executed and how their outcomes are propagated upward in the tree. This design offers two key advantages: (1) isolating subgoals reduces hallucination and logical errors in long trajectories by keeping each agent node focused on its local context and goal, which also enables targeted in-context example selection, and (2) control flow nodes ensure robust and interpretable execution logic for reliable planning.

\subsection{ReAcTree Algorithm}
\label{subsec:reactree_algorithm}
\emph{ReAcTree} algorithm dynamically constructs an agent tree of \emph{agent nodes} and \emph{control flow nodes} to accomplish a natural language goal $g$. It begins with a single agent node for the top-level goal and dynamically grows the tree as agents decide to expand. The following describes the execution of agent nodes and control flow nodes.

\textbf{Agent Nodes.} Each agent node $n$ operates as an LLM-based task planner with a specific natural language subgoal $g^n$, responsible for decision-making to achieve that goal. At each time step $t$, it accesses its context $c^n_t = (o^n_1, a^n_1, o^n_2, a^n_2, \dots, a^n_{t-1}, o^n_t)$, where $o^n_i$ and $a^n_i$ represent the observation and action at time step $i$. It then samples an action $a^n_t$ from an LLM policy $p_{\text{LLM}}(\cdot)$: 
\begin{equation}
a^n_t \sim p_{LLM}(\cdot \mid P^n, g^n, c^n_t),
\end{equation}
where the initial prompt $P^n = (P_{sys}, P^n_{ic})$ consists of a system prompt $P_{sys}$ and agent node-specific in-context examples $P^n_{ic}$.

A key feature of \emph{ReAcTree} is its extended action space, $\hat{\mathcal{A}}^n_t = \mathcal{A}^n_t \cup \mathcal{L} \cup \mathcal{E}$, where $\mathcal{A}^n_t$ represents the set of executable skills at time $t$ (e.g., \textit{pick up apple 1}); $\mathcal{L}$ is the language space for self-reasoning; and $\mathcal{E} = \mathcal{F} \times \mathcal{L}$ is the expand space, with $\mathcal{F}$ and $\mathcal{L}$ denoting the set of control flow types and the language space used to express subgoals, respectively.

If the action $a^n_t \in \mathcal{A}^n_t$ or $a^n_t \in \mathcal{L}$, the agent operates as in the \emph{ReAct} framework, either executing actions or engaging in reasoning. If $a^n_t \in \mathcal{E}$, the agent expands the tree by creating a control flow node as its child and agent nodes for the generated subgoals as grandchildren. Formally, this expansion action is expressed as $a^n_t = (f^n, [g^n_1, \dots, g^n_K])$, where $f^n$ is the control flow type and each $g^n_i$ is a natural language subgoal. A control flow node $n_f$ with type $f^n$ is then attached as a child of node $n$, and agent nodes $n_i$ with subgoals $g^n_i$ are added as children of $n_f$. After expansion, node $n$ executes the control flow node $n_f$ and awaits its result. The agent node terminates when one of the following occurs: generating \textit{done} (returning success), \textit{failure}, or reaching the maximum decision count (both returning failure).

\textbf{Control Flow Nodes.} Each control flow node coordinates the execution of its child agent nodes according to behavior tree principles \citep{colledanchise2018learning}. \emph{ReAcTree} supports three types of control flow nodes: \emph{sequence} ($\rightarrow$), \emph{fallback} ($?$), and \emph{parallel} ($\Rightarrow$). The \textit{sequence} node executes its children sequentially, returning success only if all of them succeed and failing immediately if any child fails. The \textit{fallback} node also executes children in order but returns success as soon as the first child succeeds and fails only if all children fail. Lastly, the \textit{parallel} node executes all children sequentially without early termination and aggregates outcomes according to a predefined policy; we adopt majority voting, useful for independent subgoals (e.g., placing multiple objects in different locations). After execution, a control flow node reports its overall success or failure to its parent node. The complete pseudocode is provided in Appendix~\ref{app:pseudocode}.

\subsection{Memory Systems}
\label{subsec:memory_systems}
To support \emph{ReAcTree}’s hierarchical planning and decision-making, we introduce two complementary memory systems: \emph{episodic memory} and \emph{working memory}. Episodic memory enhances the in-context learning capability of agent nodes by storing and retrieving past subgoal-level experiences. In contrast, working memory facilitates information sharing across nodes by recording environment-specific observations, such as the latest location of movable objects.

\textbf{Episodic Memory.} Each agent node $e$ handles a particular subgoal $g^e$, and its entire decision-making trajectory forms a subgoal-level experience. Episodic memory $M_{ep}$ stores experiences from agent nodes involved in task executions that ultimately succeed, even if the individual node did not complete its subgoal.

By storing experiences at the subgoal level, episodic memory maintains shorter, goal-directed trajectories than monolithic planners like \emph{ReAct}. For instance, while \emph{ReAct} stores a single long trajectory for \textit{Bring pudding and juice to the coffee table}, \emph{ReAcTree} stores focused trajectories for each subgoal, such as \textit{find and pick up pudding} and \textit{find and pick up juice}. This granularity enables more targeted in-context retrieval and improves agent decision accuracy.

Formally, each experience is recorded as a tuple $(t^e, v^e, s^e)$, where: (1) $t^e = (g^e,\,o^e_1,\,a^e_1,\,\dots,\,o^e_T,\,a^e_T)$ denotes the full text trajectory, recording all observations $o^e_t$ and actions $a^e_t$ during the node’s lifetime; (2) $v^e = f_{\text{sen}}(g^e)$ is the sentence embedding of the node’s subgoal, computed using a pretrained encoder $f_{\text{sen}}$ such as Sentence-BERT~\citep{reimers2019sentence}; and (3) $s^e \in \{\textit{success}, \textit{failure}, \textit{expand}\}$ is the termination state of the agent node.

The collected episodic memory $M_{ep}$ provides in-context examples for agent node inference. Before decision-making begins at inference time, an agent node retrieves in-context examples from $M_{ep}$ by comparing its current subgoal $g^n$ to the stored goals using cosine similarity. Specifically, the agent embeds its goal $g^n$ as $v^n = f_{\text{sen}}(g^n)$ and computes similarity with each $v^e$ in $M_{ep}$: $\text{sim}(v^n, v^e) = (v^n \cdot v^e) / (\|v^n\| \cdot  \|v^e\|)$. The agent retrieves the top $k$ similar experiences (with $k$ determined by a token limit) as in-context examples. If multiple experiences have the same similarity score, \emph{ReAcTree} samples uniformly across termination states \{\textit{success}, \textit{failure}, \textit{expand}\} to promote diversity. In practice, episodic memory can be bootstrapped from a few manually collected trajectories and gradually augmented with additional successful executions.

\textbf{Working Memory.} Working memory captures environment-specific information within a single \emph{ReAcTree} run, shared across all agent nodes to store and recall key observations. In this work, we focus on tracking movable object locations to reduce redundant interactions and mitigate hallucinations by providing accurate, environment-specific knowledge.

\begin{table*}[t!]
\centering
\caption{Performance of \emph{ReAcTree}, \emph{ReAcTree+WM}, and 5 baselines across 7 LLMs. Both GSR and SSR (\%) are reported. \textbf{Bold} indicates the best result, and \underline{underlined} indicates the second best. For brevity, we use Phi-4-RP to denote Phi-4-reasoning-plus.}
\label{tab:performance_comparison}
\begin{tabular}{l cc cc cc cc cc cc cc}
\toprule
\multirow{2}{*}{Method}
& \multicolumn{2}{c}{\shortstack{LLaMA 3.1 8B}} & \multicolumn{2}{c}{\shortstack{LLaMA 3.1 70B}} & \multicolumn{2}{c}{\shortstack{Qwen 2.5 7B}} & \multicolumn{2}{c}{\shortstack{Qwen 2.5 72B}} & \multicolumn{2}{c}{\shortstack{Mistral 7B}} & \multicolumn{2}{c}{\shortstack{Gemma 2 9B}} & \multicolumn{2}{c}{\shortstack{Phi-4-RP 14B}} \\
\cmidrule(lr){2-3} \cmidrule(lr){4-5} \cmidrule(lr){6-7} \cmidrule(lr){8-9} \cmidrule(lr){10-11} \cmidrule(lr){12-13} \cmidrule(lr){14-15}
 & GSR & SSR & GSR & SSR & GSR & SSR & GSR & SSR & GSR & SSR & GSR & SSR & GSR & SSR \\
\midrule
\emph{ZSP} & 1.00 & 13.03 & 0.00 & 14.42 & 0.00 & 8.98 & 0.00 & 14.22 & 0.00 & 11.65 & 1.00 & 13.87 & 0.00 & 17.90 \\
\emph{Tree-Planner}$_{N=25}$ & 1.00 & 17.00 & 2.00 & 16.72 & 6.00 & 22.23 & 6.00 & 32.41 & 1.00 & 20.43 & 2.00 & 17.58 & 3.00 & 17.52 \\
\emph{Tree-Planner}$_{N=50}$ & 4.00 & 21.85 & 4.00 & 23.43 & 8.00 & 28.10 & 9.00 & 36.03 & 6.00 & 23.63 & 3.00 & 23.30 & 4.00 & 20.40 \\
\emph{ReAct} & 8.00 & 34.25 & 30.00 & 57.05 & 10.00 & 31.82 & 26.00 & 51.38 & 6.00 & 28.18 & 9.00 & 37.20 & \underline{33.00} & 48.13 \\
\emph{ReAct+WM} & 16.00 & 42.65 & \underline{33.00} & \underline{63.15} & 13.00 & 39.73 & 31.00 & 54.05 & 9.00 & 31.95 & 11.00 & 39.93 & \underline{33.00} & 51.28 \\
\emph{ReAcTree} & \underline{21.00} & \underline{51.98} & 32.00 & 60.58 & \underline{18.00} & \underline{50.20} & \underline{48.00} & \underline{75.13} & \underline{11.00} & \underline{37.92} & \underline{26.00} & \underline{60.43} & \textbf{49.00} & \underline{67.47} \\
\emph{ReAcTree+WM} & \textbf{30.00} & \textbf{60.77} & \textbf{58.00} & \textbf{79.27} & \textbf{37.00} & \textbf{59.63} & \textbf{61.00} & \textbf{79.58} & \textbf{15.00} & \textbf{49.57} & \textbf{38.00} & \textbf{67.08} & \textbf{49.00} & \textbf{69.30} \\
\bottomrule
\end{tabular}
\end{table*}

Working memory is integrated into agent nodes of \emph{ReAcTree} via two mechanisms. First, the executable skill set $\mathcal{A}^n_t$ is augmented with special actions, \textit{recall location of $<$movable object$>$}, enabling agents to query the stored location of any movable object directly instead of actively exploring the environment. This recall action exclusively queries working memory. Second, working memory is automatically updated whenever an agent observes movable objects during interaction. For instance, if an agent opens a fridge and finds juice, it updates the working memory to reflect that juice is now in the fridge. We implement working memory as a lightweight Python dictionary that maps object classes to lists of observed instances and their associated locations (e.g., IDs, rooms, receptacles). This mechanism extends the concept of tool usage in language models \citep{schick2024toolformer}, treating \textit{recall location} as a specialized tool-like action.

%% file: latex/05_experiments.tex
\section{Experiments}
\label{experiments}

\subsection{Experimental Setup}
\label{subsec:experimental_setup}
\textbf{Datasets and Simulators.}
We follow the evaluation protocol of LoTa-Bench~\citep{choi2024lotabench}, which provides two dataset-simulator pairs for language-based task planning: WAH-NL~\citep{puig2021watchandhelp, choi2024lotabench} with VirtualHome~\citep{puig2018virtualhome}, and ALFRED~\citep{shridhar2020alfred} with AI2THOR~\citep{kolve2017ai2}. We extend both environments to more realistic, partially observable settings, where the agent receives limited textual observations after each action. All objects additionally include class and instance identifiers for precise grounding. We conduct primary experiments on WAH-NL with VirtualHome, as it features longer-horizon, multi-room tasks with multiple subgoals, and use ALFRED with AI2THOR, which contains relatively short-horizon, single-room, single-goal tasks, mainly for complementary validation.

WAH-NL, an extension of WAH~\citep{puig2021watchandhelp}, comprises 250 training and 100 test tasks across five categories. We use the training set to build episodic memory and the test set for evaluation. To improve reliability, we manually corrected four test tasks with mismatched instructions and goal conditions (see Appendix \ref{app:testset_revision} for details). ALFRED includes seven task types. In LoTa-Bench, the \textit{pick and place two objects} task was omitted due to the absence of instance identifiers. We include this task in our evaluation by incorporating simulator-provided instance identifiers. As with WAH-NL, the training set is used to build episodic memory, and evaluation is conducted on the valid-seen and valid-unseen splits.

In both simulators, the agent executes natural language-based primitive actions. After each action, it receives an action-dependent textual observation generated from the simulator’s ground-truth state, revealing only visible objects and receptacles. VirtualHome supports six primitive actions (\textit{go to, pick up, put down, open, close, turn on}), while AI2THOR supports eight (\textit{go to, pick up, put down, open, close, turn on, turn off, slice}). Examples of action-observation pairs are provided in Appendix~\ref{app:exp_settings}.

\textbf{Evaluation Metrics.}
We evaluate task planning performance using two metrics. The \textit{goal success rate} (GSR) is the percentage of tasks where the agent successfully achieves the overall goal. The \textit{subgoal success rate} (SSR) is the ratio of completed subgoals to the total number of subgoals. We report both GSR and SSR for WAH-NL, and only GSR for ALFRED, which lacks explicit subgoal definitions.

\textbf{Implementation Details.}
We evaluate \emph{ReAcTree} and its working memory variant, \emph{ReAcTree+WM}, in a few-shot in-context learning setting without fine-tuning. By default, all experiments include episodic memory unless otherwise specified.

To bootstrap episodic memory, we first manually collected a small number of trajectories and used them as in-context examples to run \emph{ReAcTree} with LLaMA-3.1 70B~\citep{dubey2024llama} over the training set. Only successful executions were retained in episodic memory. For WAH-NL, we collected one manual trajectory per task type and executed the full training set. For ALFRED, we collected three per type and stored up to 100 successful trajectories per type to reduce computational overhead, given the large training set (21K tasks). 

The retrieval size was constrained such that the total length of retrieved trajectories used as in-context examples did not exceed 5K tokens. The decision cap was set to 200 for WAH-NL and 100 for ALFRED. We used the Guidance library~\citep{guidance} for text generation, employing temperature 0.0 for free-form generation and leveraging its constrained generation capabilities to deterministically select acting actions and control flow types. Prompt templates for both WAH-NL and ALFRED are available in Appendix~\ref{app:reactree_prompts}.

\textbf{Baselines.}
We compare \emph{ReAcTree} and \emph{ReAcTree+WM} against five baselines: \emph{ZSP} \citep{huang2022language}, \emph{Tree-Planner} \citep{hu2024treeplanner} with $N=25$ or $50$ sampled plans, \emph{ReAct} \citep{yao2023react}, and its working memory variant \emph{ReAct+WM}. All baselines are evaluated on WAH-NL using the same retrieval size and decision cap as \emph{ReAcTree}. For ALFRED, we report results only for \emph{ReAct+WM}, as it serves as the strongest baseline on WAH-NL. Appendices~\ref{app:baselines_details} and \ref{app:prompts} provide implementation details and prompts. 

\begin{figure*}[t]
    \centering
    \begin{minipage}{\textwidth}
        \centering
        \includegraphics[width=0.9\textwidth]{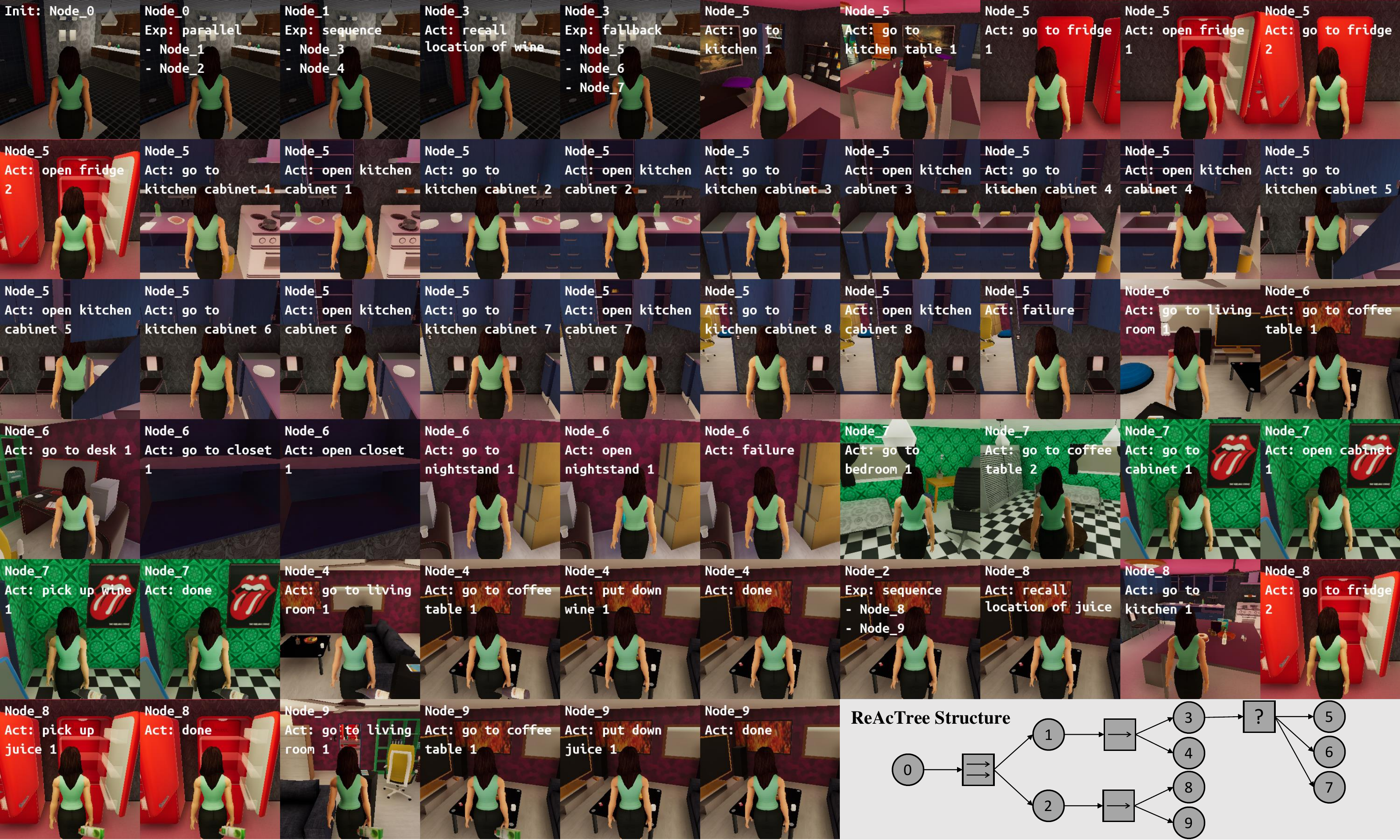}
    \end{minipage}
    \begin{minipage}{\textwidth}
        \centering
        {\begin{adjustbox}{width=0.9\textwidth}
        \begin{tabular}{ll||ll}
            Node 0 & \textit{Ensure that both a wine and a juice are on the coffee table.} & Node 5 & \textit{Find and pick up the wine in kitchen 1.} \\
            Node 1 & \textit{Move the wine onto the coffee table.} & Node 6 & \textit{Find and pick up the wine in living room 1.}\\
            Node 2 & \textit{Move the juice onto the coffee table.} & Node 7 & \textit{Find and pick up the wine in bedroom 1.}\\
            Node 3 & \textit{Find and pick up the wine.} & Node 8 & \textit{Find and pick up the juice.} \\
            Node 4 & \textit{Find the coffee table and put down the wine.} & Node 9 & \textit{Find the coffee table and put down the juice.} \\
        \end{tabular}
        \end{adjustbox}}
    \end{minipage}
    \caption{Success case of \emph{ReAcTree} on the WAH-NL dataset using Qwen~2.5~72B. The snapshot of each step and the tree structure of nodes are shown. The subgoal of each node is also listed.}
    \label{fig:reactree_success_ex}
    \Description{A success case of ReAcTree in the WAH-NL simulator. The figure displays step-by-step agent actions for a complex task, accompanied by the generated behavior tree structure and a table listing the specific subgoal for each node.}
\end{figure*}

\subsection{Main Results}
We first present the primary evaluation results on the WAH-NL dataset, which features complex, long-horizon tasks under partial observability. Table~\ref{tab:performance_comparison} compares five baselines (\emph{ZSP}, \emph{Tree-Planner} with $N=25$ or $50$, \emph{ReAct}, \emph{ReAct+WM}) and our methods (\emph{ReAcTree}, \emph{ReAcTree+WM}) across seven LLMs ranging from small (7--14B) to large (70--72B), including LLaMA 3.1~\citep{dubey2024llama}, Qwen 2.5~\citep{yang2024qwen2}, Mistral~\citep{jiang2023mistral}, Gemma 2~\citep{team2024gemma}, and Phi-4-reasoning-plus~\citep{abdin2025phi} (see Appendix~\ref{app:lang_models} for the full list). Both GSR and SSR are reported.

\emph{ZSP} generates a complete plan at once and thus struggles to adapt under partial observability, resulting in poor performance. \emph{Tree-Planner} improves this by sampling multiple high-level plans ($N=25$ or $50$) to construct an action tree and adapting its actions based on observations. However, if none of the sampled plans succeed, the system cannot recover. This limitation becomes critical as the search space grows. For example, a typical kitchen may contain multiple \textit{kitchen tables}, \textit{kitchen counters}, \textit{fridges}, and a \textit{dishwasher}. Without knowing in advance which receptacle contains the target object, plan sampling becomes intractable and frequently fails to generate a valid plan. Even after relaxing action constraints for \textit{pick up} and \textit{put down} by omitting the instance identifier (e.g., \textit{pick up apple} instead of \textit{pick up apple 1}), the method still yields only marginal improvements over \emph{ZSP} in our partially observable setting.

Both \emph{ReAct} and \emph{ReAcTree} significantly improve GSR and SSR over \emph{ZSP} and \emph{Tree-Planner}. While \emph{ReAct} shows limited gains, \emph{ReAcTree} consistently achieves higher scores across all LLMs. For example, with Qwen 2.5 72B, \emph{ReAcTree+WM} achieves a GSR of 61\% compared to 31\% for \emph{ReAct+WM}. Its SSR also rises from 54.05\% to 79.58\%. These consistent gains highlight the effectiveness of decomposing complex tasks into manageable subgoals, where agent nodes handle each semantically isolated subgoal within a focused context and control flow nodes coordinate execution strategies, together making long-horizon tasks more tractable under partial observability.

Notably, \emph{ReAcTree} enables smaller LLMs to outperform baselines with much larger models. For example, \emph{ReAcTree+WM} with Qwen 2.5 7B surpasses all baselines in GSR and most in SSR, even against Qwen 2.5 72B and LLaMA 3.1 70B. This improvement stems from two factors: (1) decomposing tasks into simpler subgoals, which keeps the accumulating decision-making trajectory focused on one subgoal at a time, and (2) retrieving subgoal-level examples that are more directly comparable than task-level retrieval. Together, these factors enable smaller models to reason more effectively, substantially narrowing the performance gap across model scales.

To demonstrate how \emph{ReAcTree} plans and executes complex tasks, we analyze a representative task: \textit{Make sure there is a wine and a juice on the coffee table.} As shown in Figure~\ref{fig:reactree_success_ex}, \emph{ReAcTree+WM} successfully solves it by decomposing into two subgoals: \textit{move the wine onto the coffee table} and \textit{move the juice onto the coffee table}, and executing them in parallel. It explores multiple rooms using fallback strategies, whereas \emph{ReAct+WM} remains confined to \textit{kitchen 1}, failing to locate the wine in \textit{bedroom 1}. Full trajectories are in Appendix~\ref{app:success_traj_reactree_wm}, and the corresponding \emph{ReAct+WM} failure case is visualized in Appendix~\ref{app:react_fail_ex}.

\begin{table}[t!]
\centering
\setlength{\tabcolsep}{3pt}
\caption{Performance of \emph{ReAct+WM} and \emph{ReAcTree+WM} on the ALFRED across 3 LLMs. GSR (\%) is reported for both valid-seen and valid-unseen splits. \textbf{Bold} indicates the best result.}
\label{tab:alfred_split_grouped}
\begin{tabular}{ll cccc c}
\toprule
\multirow{2}{*}{Split} & \multirow{2}{*}{Method} & \multicolumn{2}{c}{LLaMA 3.1} & \multicolumn{2}{c}{Qwen 2.5} & \multicolumn{1}{c}{Phi-4-RP} \\
\cmidrule(lr){3-4} \cmidrule(lr){5-6} \cmidrule(lr){7-7}
& & 8B & 70B & 7B & 72B & 14B \\
\midrule
\multirow{2}{*}{\textit{\shortstack{Valid-\\Seen}}} & \emph{ReAct+WM} & 21.22 & 33.31 & 16.83 & 37.07 & 31.71 \\
& \emph{ReAcTree+WM} & \textbf{25.85} & \textbf{40.00} & \textbf{20.98} & \textbf{40.85} & \textbf{35.12} \\
\midrule
\multirow{2}{*}{\textit{\shortstack{Valid-\\Unseen}}} & \emph{ReAct+WM} & 19.61 & 32.40 & 16.81 & 39.10 & 29.72 \\
& \emph{ReAcTree+WM} & \textbf{26.19} & \textbf{37.03} & \textbf{25.09} & \textbf{39.83} & \textbf{36.18} \\
\bottomrule
\end{tabular}
\end{table}

\begin{table*}[t!]
\caption{Memory ablation results on Qwen 2.5 7B and 72B. \textbf{EM, WM} describes the memory configuration: none (\xmark, \xmark), WM only (\xmark, \checkmark), EM only (\checkmark, \xmark), or EM+WM (\checkmark, \checkmark). We report GSR (\%) and SSR (\%) with \(\Delta\) improvements over the no-memory baseline.}
\label{tab:mem_ablation_full}
\begin{tabular}{@{}lccccc@{}}
\toprule
\multirow{2}{*}{\textbf{Method}} & \multirow{2}{*}{\textbf{EM, WM}} & 
\multicolumn{2}{c}{Qwen 2.5 7B} & 
\multicolumn{2}{c}{Qwen 2.5 72B} \\
\cmidrule(lr){3-4} \cmidrule(lr){5-6}
 &  & \textbf{GSR ($\Delta$)} & \textbf{SSR ($\Delta$)} & \textbf{GSR ($\Delta$)} & \textbf{SSR ($\Delta$)} \\
\midrule

\multirow{4}{*}{\emph{ReAct}}           
    & \xmark, \xmark        & \ph7.00\hspace{3.0em}   & 22.82\hspace{3.4em}  & 13.00\hspace{3.4em}  & 35.75\hspace{3.4em} \\
    & \xmark, \checkmark    & \ph6.00 (--1.00)        & 19.53 (--\ph3.29)    & 18.00 (+\ph5.00)    & 44.33 (+\ph8.58) \\
    & \checkmark, \xmark    & 10.00 (+3.00)        & 31.82 (+\ph9.00)     & 26.00 (+13.00)      & 51.38 (+15.63) \\
    & \checkmark, \checkmark& 13.00 (+6.00)        & 39.73 (+16.91)       & 31.00 (+18.00)      & 54.05 (+18.30) \\
\midrule 
\multirow{4}{*}{\emph{ReAcTree}}        
    & \xmark, \xmark        & \ph2.00\hspace{3.4em}   & \ph9.32\hspace{3.4em} & 31.00\hspace{3.4em}  & 56.82\hspace{3.4em} \\
    & \xmark, \checkmark    & \ph1.00 (--\ph1.00)     & \ph7.45 (--\ph1.87)   & 47.00 (+16.00)      & 64.72 (+\ph7.90) \\
    & \checkmark, \xmark    & 18.00 (+16.00)          & 50.20 (+40.88)        & 48.00 (+17.00)      & 75.13 (+18.31) \\
    & \checkmark, \checkmark& 37.00 (+35.00)          & 59.63 (+50.31)        & 61.00 (+30.00)      & 79.58 (+22.76) \\
\bottomrule
\end{tabular}
\end{table*}

We further evaluate \emph{ReAcTree+WM} on the ALFRED dataset. As shown in Table~\ref{tab:alfred_split_grouped}, it consistently outperforms \emph{ReAct+WM} across all evaluated models and splits. For instance, on the valid-unseen split, it improves over \emph{ReAct+WM} by 6.58 and 4.63 percentage points with LLaMA 3.1 8B and 70B, respectively, with similar gains observed for Qwen 2.5 and Phi-4-reasoning-plus. These results indicate that \emph{ReAcTree} remains effective even in relatively short-horizon, single-room tasks and generalizes well to unseen environments.

Beyond quantitative gains, \emph{ReAcTree+WM} also handles complex tasks requiring more precise procedures. For instance, given the instruction \textit{place a cooked potato slice in the fridge}, both methods slice the potato, but \emph{ReAct+WM} skips the cooking step and places it directly in the fridge, whereas \emph{ReAcTree+WM} correctly heats it using the microwave before storage by decomposing the instruction into four subgoals with a sequence control flow node: \textit{find and pick up knife}, \textit{slice and pick up potato}, \textit{cook and pick up potato}, and \textit{place potato in fridge} (see Appendix~\ref{app:qualitative_results_for_alfred} for trajectories of both methods).

\subsection{Memory Ablation}
All subsequent experiments in this section are conducted on the WAH-NL dataset, featuring long-horizon, multi-room tasks. We study the impact of \emph{episodic memory (EM)} and \emph{working memory (WM)} through ablations with Qwen 2.5 models (7B and 72B). When EM is disabled, we use the manually annotated trajectory of a randomly selected training task as a fixed in-context example for all test tasks. For \emph{ReAcTree}, trajectories from all agent nodes within the selected task are concatenated into a single in-context example. Table~\ref{tab:mem_ablation_full} reports GSR and SSR of \emph{ReAct} and \emph{ReAcTree} under four configurations: no memory, WM only, EM only, and EM+WM.

\textbf{EM and WM show strong synergy.} As shown in Table~\ref{tab:mem_ablation_full}, adding either memory component generally improves performance, but combining them consistently yields the highest performance, underscoring their complementary roles. EM provides clean, semantically relevant in-context examples, while WM retains task-relevant observations.

\textbf{Model scale is a critical factor in EM-disabled settings.} The results also highlight a nuanced interaction between memory and model scale, especially when EM is disabled. On the 7B model, both \emph{ReAct} and \emph{ReAcTree} perform worse in the WM-only setting than the no-memory baseline. This counterintuitive result occurs because, without EM, the agent receives a fixed and potentially mismatched in-context example. Adding WM further introduces complexity (the \texttt{recall location of <object>} skill) that is not well-grounded by the poor example, leading to confusion for the smaller model with its limited reasoning capacity. 

In contrast, the 72B model is robust to this degradation. Its performance improves significantly with WM alone (+16\%p GSR for \emph{ReAcTree}), as its greater reasoning capacity handles the additional complexity gracefully, even with suboptimal in-context examples. 

This dynamic also explains why performance between \emph{ReAcTree} and \emph{ReAct} reverses with scale. When EM is disabled, the 7B \emph{ReAcTree} underperforms because it struggles to parse the complex, multi-granular in-context example, whereas \emph{ReAct}'s simpler, flat prompt structure is easier for the 7B model to follow. However, at the 72B scale, this trend reverses dramatically: the larger model has sufficient capacity to leverage the hierarchical structure, allowing \emph{ReAcTree} to outperform \emph{ReAct} by a large margin (e.g., +18\%p in the no-memory setting and +29\%p in the WM-only setting).

\textbf{Hierarchical decomposition is fundamentally powerful.} This analysis shows that the hierarchical structure of \emph{ReAcTree} provides inherent value. The most compelling evidence is that even without memory support, the 72B \emph{ReAcTree} achieves a GSR of 31.00\%, substantially outperforming the 72B \emph{ReAct} at 13.00\%. This confirms that the benefits of hierarchical decomposition are fundamental to the architecture, independent of memory components.

\begin{table}[t!]
\centering
\setlength{\tabcolsep}{3pt} 
\caption{Control flow ablation results. GSR and SSR (\%) are reported. \textbf{Bold} indicates the best performance.}
\label{tab:control_flow_ablation}
\begin{tabular}{l c c c c c c c c}
\toprule
\multirow{2}{*}{\shortstack{Ctrl \\ Flow}} 
 & \multicolumn{2}{c}{\shortstack{LLaMA 3.1 \\ 8B}} 
 & \multicolumn{2}{c}{\shortstack{LLaMA 3.1 \\ 70B}} 
 & \multicolumn{2}{c}{\shortstack{Qwen 2.5 \\ 7B}} 
 & \multicolumn{2}{c}{\shortstack{Qwen 2.5 \\ 72B}} \\
\cmidrule(lr){2-3}
\cmidrule(lr){4-5}
\cmidrule(lr){6-7}
\cmidrule(lr){8-9}
 & GSR & SSR & GSR & SSR & GSR & SSR & GSR & SSR \\
\midrule
\textit{all}     & \textbf{30.00} & \textbf{60.77} & \textbf{58.00} & \textbf{79.27} & 37.00 & \textbf{59.63} & \textbf{61.00} & \textbf{79.58} \\
\textit{seq+fb}  & 28.00 & 58.77 & \textbf{58.00} & 78.52 & \textbf{38.00} & 54.83 & \textbf{61.00} & 79.08 \\
\textit{seq}     & 18.00 & 45.65 & 36.00 & 61.17 & 15.00 & 32.18 & 46.00 & 63.22 \\
\bottomrule
\end{tabular}
\end{table}

\subsection{Control Flow Ablation}
We conducted a control flow ablation study of \emph{ReAcTree} with three configurations: all control flows (\textit{all}), sequence and fallback (\textit{seq+fb}), and sequence only (\textit{seq}). Following the same experimental protocol, we manually collected one trajectory per task type for each setting and bootstrapped additional successful trajectories with LLaMA 3.1 70B to construct episodic memory.

As shown in Table~\ref{tab:control_flow_ablation}, using \textit{all} consistently yields the best performance, while \textit{seq+fb} performs comparably. In contrast, relying solely on \textit{seq} significantly degrades performance. These results highlight the importance of expressive control flows, which coordinate dependencies, enable parallel execution, and recover from subgoal failures in long-horizon tasks.

\subsection{Computational Cost Analysis}
\label{subsec:computational_cost_analysis}

To analyze computational cost of \emph{ReAcTree}, we compared three representative configurations using LLaMA 3.1: \emph{ReAct+WM} (70B), and \emph{ReAcTree+WM} (70B, 8B). All methods were evaluated on the same 19 tasks successfully completed by all configurations. For fairness, all experiments were run on identical hardware with two H100 GPUs and one H200 GPU. Our analysis considers two perspectives: (1) the trade-off between performance and efficiency, and (2) token-based resource usage as a proxy for GPU memory demand.

First, Table \ref{tab:exec_time} compares the execution time, number of decision steps, and overall performance. At the same 70B model size, \emph{ReAcTree+WM} requires more decision steps and a longer runtime but achieves a significantly higher GSR (+25\%p) than \emph{ReAct+WM}. This runtime difference stems from a greater number of decision steps rather than a higher computational cost per decision, representing a trade-off for increased robustness and success rate. Notably, \emph{ReAcTree+WM} (8B) attains performance comparable to \emph{ReAct+WM} (70B) while being significantly faster, highlighting its efficiency in low-resource settings.

\begin{table}[t!]
\centering
\setlength{\tabcolsep}{4pt}
\caption{Average execution time, decision steps, and GSR/SSR on shared successful tasks.}
\label{tab:exec_time}
\begin{tabular}{@{}lccc@{}}
\toprule
\textbf{Configuration} & \textbf{Time (s)} & \textbf{Decision Steps} & \textbf{GSR/SSR (\%)} \\
\midrule
\emph{ReAct+WM} (70B)    & 109.1 & 60.1 & 33 / 62.15  \\
\emph{ReAcTree+WM} (70B) & 198.6 & 75.2 & 58 / 79.27  \\
\emph{ReAcTree+WM} (8B)  & 69.9  & 78.0 & 30 / 60.77  \\
\bottomrule
\end{tabular}
\end{table}

To measure computational resource requirements, we analyzed token usage. Specifically, the maximum number of input tokens serves as an indicator of peak GPU memory demand, while the average number of input/output tokens per decision step reflects the per-step computational load. The results are summarized in Table \ref{tab:token_usage}. The average token usage per decision is similar across all models, confirming that the tested configurations of \emph{ReAcTree} and \emph{ReAct} incur a comparable computational cost per step. However, \emph{ReAct+WM} (70B) shows the highest peak input token count (8316) and the largest variance, suggesting greater GPU memory overhead and less predictable resource usage. In contrast, \emph{ReAcTree} maintains well-bounded and stable token usage due to its modular structure, where each agent node processes a localized subgoal.

\begin{table}[t!]
\centering
\setlength{\tabcolsep}{2pt}
\caption{Peak and average token usage per decision.}
\label{tab:token_usage}
\begin{tabular}{@{}lccc@{}}
\toprule
\textbf{Configuration} & \textbf{Max Input} & \textbf{Avg. Input} & \textbf{Avg. Output}  \\
\midrule
\emph{ReAct+WM} (70B)    & 8316 & 5359.45 (± 904.83) & 16.07 (± 1.95) \\
\emph{ReAcTree+WM} (70B) & 6977 & 5362.66 (± 109.06) & 17.83 (± 1.50) \\
\emph{ReAcTree+WM} (8B)  & 7173 & 5390.12 (± 125.72) & 17.68 (± 1.73) \\
\bottomrule
\end{tabular}
\end{table}

\subsection{Failure Case Analysis}
\label{subsec:failure_case_analysis}
We analyzed 39 failure cases of \emph{ReAcTree+WM} with Qwen 2.5 72B in WAH-NL, categorizing them into four types: \textit{Ambiguous} (10), \textit{Execution} (12), \textit{Search} (13), and \textit{Expand} (4), as visualized in Figure~\ref{fig:failure_cases}.

\textit{Ambiguous} cases arose from vague instructions (e.g., \textit{give me 2 drinks}), creating uncertainty about object types or locations. \textit{Execution} failures, largely due to LLM hallucination, included object confusion (\textit{Confusion}, 8), irrelevant action repetition (\textit{Repeat}, 2), and overlooked targets (\textit{Skipping}, 2), reflecting challenges in reasoning consistency and object recognition. \textit{Search} failures, the most frequent, resulted from exploration limitations under partial observability, including incomplete searches (\textit{PlanFail}, 6), local search loops (\textit{SameRm}, 3), and unnecessary revisits (\textit{Revisit}, 2). Finally, \textit{Expand} failures involved incorrect subgoal decomposition, with missing (\textit{Missing}, 2) or incorrect (\textit{Wrong}, 2) subgoals.

\textit{Search} failures were dominant, underscoring the need for stronger fallback and exploration strategies. Addressing \textit{Ambiguous} cases requires clarifying dialogues, while \textit{Execution} errors highlight the importance of hallucination control. Less frequent \textit{Expand} failures point to the need for subgoal refinement in complex scenarios.

%% file: latex/06_conclusion.tex
\begin{figure}[t!]
    \centering
    \includegraphics[width=0.38\textwidth]{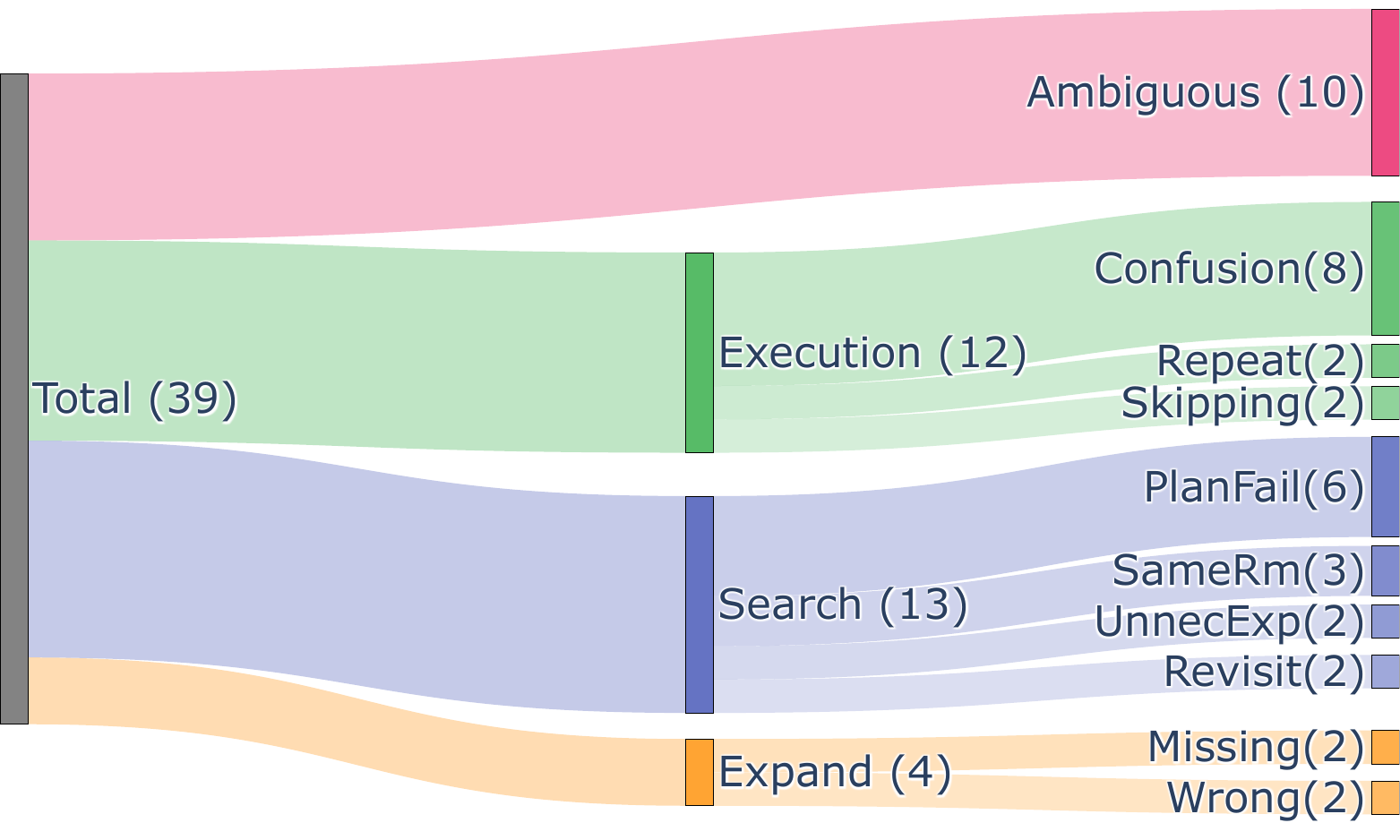}
    \caption{Categories and subcategories of failure cases for \emph{ReAcTree+WM}.}
    \label{fig:failure_cases}
    \Description{A Sankey diagram showing the breakdown of failure cases into four main categories (Search, Execution, Ambiguous, Expand) and their respective sub-types.}
\end{figure}

\section{Conclusion}
\label{conclusion}
We introduced \emph{ReAcTree}, a hierarchical task planner that extends \emph{ReAct} by dynamically constructing an agent tree in subgoal space. Each agent node can reason, act, or expand the tree with new subgoals, while control flow nodes coordinate their execution through sequence, fallback, or parallel strategies. This design prevents error accumulation typical of single-pass trajectories and makes long-horizon tasks more tractable. To strengthen in-context learning and coordination, we introduced episodic memory, storing subgoal-level experiences for retrieval, and working memory, sharing environment-specific observations across nodes. Experiments on WAH-NL and ALFRED under partial observability show that \emph{ReAcTree} consistently outperforms baselines across diverse LLMs, demonstrating improved reliability, generalization to unseen environments, and even enabling smaller models to rival larger ones.

Despite these benefits, \emph{ReAcTree} still faces LLM-specific challenges, including hallucinations and limited capability to recognize failures. It also lacks mechanisms to revise incorrectly expanded subgoals and to handle instruction ambiguity. Future work will focus on mitigating hallucination, refining subgoal correction, and enabling clarification dialogues, ultimately advancing the robustness and deployability of \emph{ReAcTree} in complex, real-world environments.

%% file: appendix.tex
\section{ReAcTree Pseudocode}
\label{app:pseudocode}
This section presents the pseudocode for the two core components of \emph{ReAcTree}: agent node execution (Algorithm~\ref{alg:exec_agent_node}) and control flow node execution (Algorithm~\ref{alg:exec_ctrlflow_node}). The planning process begins with a single agent node $n_0$ assigned to the top-level goal $g$, which is executed through a call to \textsc{ExecAgentNode}($n_0$, $D_{init}=0$), where $D_{init}$ denotes the initial decision count.

At each step, an agent node samples an action from the LLM, selecting from reasoning, acting, and expanding. When expanding, the agent node creates a control flow node as its child. This control flow node then instantiates agent nodes for the generated subgoals as its own children and executes them according to its behavior type, reporting success or failure back to its parent. Algorithms~\ref{alg:exec_agent_node} and \ref{alg:exec_ctrlflow_node} summarize these procedures.

\begin{algorithm*}[h!]
\caption{\textsc{ExecAgentNode}($n$, $D_{init}$): Execution of an agent node}
\label{alg:exec_agent_node}
\begin{algorithmic}[1]
\State \textbf{Input:} Agent node $n$ with subgoal $g^n$, initial decision count $D_{\text{init}}$, maximum decision count $D_{\max}$
\State $D \gets D_{\text{init}}$, $t \gets 1$, \textit{terminate} $\gets$ \textbf{False}
\State $P^n_{\text{ic}} \gets \textsc{RetrieveFromEM}(g^n)$ \Comment{Retrieve in-context examples from episodic memory}
\State $P^n \gets (P_{\text{sys}}, P^n_{\text{ic}})$ \Comment{Initialize LLM prompt}
\State $c^n_t \gets (o^n_t)$ \Comment{Initialize context with current observation}
\While{\textbf{not} terminate}
    \State $a^n_t \sim p_{\text{LLM}}(\cdot \mid P^n, g^n, c^n_t)$ \Comment{Sample next action from LLM}
    \State $D \gets D + 1$
    \State $t \gets t + 1$
    \If{$a^n_t = \textit{done}$}
        \State \Return \textit{Success, D}
    \ElsIf{$a^n_t = \textit{failure}$ \textbf{or} $D \ge D_{\max}$}
        \State \Return \textit{Failure, D}
    \ElsIf{$a^n_t \in \mathcal{A}^n_t$} \Comment{Acting: execute an action and receive a textual observation}
        \State $o^n_{t+1} \gets \textsc{Interact}(a^n_t)$
        \State $c^n_{t+1} \gets \textsc{Append}(c^n_t, a^n_t, o^n_{t+1})$
    \ElsIf{$a^n_t \in \mathcal{L}$} \Comment{Reasoning: produce an internal thought step}
        \State $o^n_{t+1} \gets \varnothing$
        \State $c^n_{t+1} \gets \textsc{Append}(c^n_t, a^n_t, o^n_{t+1})$
    \ElsIf{$a^n_t \in \mathcal{E}$} \Comment{Expanding: generate control flow and subgoals}
        \State $(f^n, [g^n_1, \dots, g^n_K]) \gets \textsc{ParseExpand}(a^n_t)$
        \State $n_f \gets \textsc{InitCtrlFlowNode}(f^n)$
        \State \textsc{AddChild}($n$, $n_f$)
        \For{$i = 1$ to $K$}
            \State $n_i \gets \textsc{InitAgentNode}(g^n_i)$
            \State \textsc{AddChild}($n_f$, $n_i$)
        \EndFor
        \State \Return \textsc{ExecCtrlFlowNode}($n_f$, D) \Comment{Execute control flow node (Alg.~\ref{alg:exec_ctrlflow_node})}
    \EndIf
\EndWhile
\end{algorithmic}
\end{algorithm*}
\clearpage

\begin{algorithm*}[h!]
\caption{\textsc{ExecCtrlFlowNode}($n_f$, $D_{init}$): Execution of a control flow node}
\label{alg:exec_ctrlflow_node}
\begin{algorithmic}[1]
\State \textbf{Input:} Control flow node $n_f$ with type $f^n$ and child agent nodes $\{n_i\}$, initial decision count $D_{\text{init}}$
\State $D \gets D_{\text{init}}$
\If{$f^n = \textit{sequence}$}
    \For{each child node $n_i$ in order}
        \State $\mathit{status}, D \gets \textsc{ExecAgentNode}(n_i, D)$ \Comment{Execute agent node (Alg.~\ref{alg:exec_agent_node})}
        \If{$\mathit{status} = \textit{Failure}$}
            \State \Return \textit{Failure}, D
        \EndIf
    \EndFor
    \State \Return \textit{Success}, D
\ElsIf{$f^n = \textit{fallback}$}
    \For{each child node $n_i$ in order}
        \State $\mathit{status}, D \gets \textsc{ExecAgentNode}(n_i, D)$ \Comment{Execute agent node (Alg.~\ref{alg:exec_agent_node})}
        \If{$\mathit{status} = \textit{Success}$}
            \State \Return \textit{Success}, D
        \EndIf
    \EndFor
    \State \Return \textit{Failure}, D
\ElsIf{$f^n = \textit{parallel}$}
    \State $\mathit{statuses} \gets [\,]$
    \For{each child node $n_i$ in order}
        \State $\mathit{status}_i, D \gets \textsc{ExecAgentNode}(n_i, D)$ \Comment{Execute agent node (Alg.~\ref{alg:exec_agent_node})}
        \State $\mathit{statuses}.\text{append}(\mathit{status}_i)$
    \EndFor
    \State $\textit{status} \gets \textsc{AggregatePolicy}(\mathit{statuses})$
    \State \Return $\textit{status}$, D
\EndIf
\end{algorithmic}
\end{algorithm*}

\section{Implementation Details for Experimental Setup}
\subsection{WAH-NL Test Set Corrections}
\label{app:testset_revision}
We identified four WAH-NL test tasks where the natural language instructions were inconsistent with the goal conditions, such as referencing incorrect object classes or omitting required items. Therefore, to ensure reliable evaluation, we revised these instructions to better align with their intended goals. Table~\ref{tab:testset-corrections} presents the goal conditions, original instructions, and revised versions.

\begin{table*}[h!]
\centering
\caption{Summary of corrections four WAH-NL test tasks were misaligned with their goal conditions.}
\begin{tabular}{p{1.2cm} p{5cm} p{4cm} p{4cm}}
\toprule
\textbf{Task ID} & \textbf{Goal Condition} & \textbf{Original Instruction} & \textbf{Revised Instruction} \\
\midrule
4 & \{`on\_juice\_coffeetable':1, `on\_wine\_coffeetable':1\} & 
Make sure there is a wine and a juice on the kitchen table. & 
Make sure there is a wine and a juice on the coffee table.\\
\midrule
8 & \{`on\_wine\_coffeetable':1, `on\_apple\_coffeetable':1\} & 
Can you please put the apple and wine that are on the coffee table into the fridge please. & 
Can you please put the apple and the wine on the coffee table? \\
\midrule
31 & \{`inside\_waterglass\_dishwasher':1, `inside\_wineglass\_dishwasher':1, `inside\_cutleryfork\_dishwasher':1, `turnOn\_dishwasher':1\} & 
Please, put the cutlery fork and the wine glass in the dishwasher and then turn it on. & 
Please put the cutlery fork, the wine glass, and the water glass in the dishwasher and then turn it on. \\
\midrule
94 & \{`on\_plate\_kitchentable':1, `on\_waterglass\_kitchentable':1, `on\_wineglass\_kitchentable':1, `on\_cutleryfork\_kitchentable':1\} & 
Please, put 1 wine glass, 1 water glass, and 1 plate on the table & 
Please put 1 wine glass, 1 water glass, 1 plate, and 1 cutlery fork on the table \\
\bottomrule
\end{tabular}
\label{tab:testset-corrections}
\end{table*}

\subsection{Partially Observable Settings}
\label{app:exp_settings}
We extended LoTa-Bench~[10] by implementing a rule-based observation generator for both VirtualHome and AI2THOR to support the partially observable setting. After each action, the agent receives a textual description of visible objects and receptacles in its current room, excluding items hidden inside closed containers. These observations are derived from the simulator’s ground-truth state and include class and instance identifiers to enable precise object grounding.

\textbf{VirtualHome.}
In VirtualHome, we apply the partially observable protocol, which supports six natural language-based primitive actions (\textit{go to}, \textit{pick up}, \textit{put down}, \textit{open}, \textit{close}, and \textit{turn on}). To capture the variability in observations, we define eight action-dependent observation rules, each specifying how textual feedback is generated based on the semantics of the executed action. Table~\ref{tab:vh_obs_ex} presents representative action-observation pairs illustrating these rules.

\textbf{AI2THOR.}
We apply the same protocol to AI2THOR, which supports eight primitive actions: \textit{go to}, \textit{pick up}, \textit{put down}, \textit{slice}, \textit{open}, \textit{close}, \textit{turn on}, and \textit{turn off}. Table~\ref{tab:thor_obs_ex} presents representative action-observation pairs in AI2THOR, illustrating how textual feedback differs depending on the executed action.

\begin{table*}[h!]
\caption{Representative action-observation pairs in VirtualHome under the partially observable setting.}
\label{tab:vh_obs_ex}
\centering
\small
\begin{tabularx}{\textwidth}{@{}p{2.3cm}p{2.6cm}X@{}}
\toprule
\textbf{Action Context} & \textbf{Action} & \textbf{Text Observation} \\
\midrule
\textit{initialization} & - & You are in the house, and there are 4 rooms: bathroom (1), bedroom (1), kitchen (1), living room (1). You are in the middle of a bathroom (1). Looking quickly around the room, you see bathroom cabinet (1), bathroom counter (1), faucet (1), sink (1), toilet (1), towel rack (1), washing machine (1). \\
\addlinespace
\textit{go to room} & go to kitchen 1 & You move to the kitchen (1). Looking quickly around the room, you see chair (1, 2, 3, 4), dishwasher (1), faucet (1), fridge (1), kitchen cabinet (1, 2, 3, 4, 5, 6, 7, 8), kitchen counter (1, 3), kitchen table (1), microwave oven (1), nightstand (1, 2), sink (1), sofa (1), stove (1), toaster (1). \\
\addlinespace
\textit{go to object or receptacle} & go to kitchen table 1 & You arrive at the kitchen table (1). You see apple (2), book (1), bowl (2, 3), cereal (1), chair (1, 2, 3, 4), chocolate syrup (1), condiment bottle (1, 2), condiment shaker (1, 2), cupcake (1, 2), cutlery fork (2, 3, 4), cutlery knife (3, 5, 6), juice (1), kitchen table (1), pancake (1), pear (1), plate (2, 3, 4, 5), pudding (2), water glass (1, 2, 4) \\
\addlinespace
\textit{pick up object} & pick up juice 1 & You pick up juice. You hold juice (1). \\
\addlinespace
\textit{put down object} & put down apple 1 & You put down apple on coffee table. \\
\addlinespace
\textit{open receptacle} & open fridge 1 & You open fridge. You see bench (1), cupcake (1), fridge (1, 2), juice (1). \\
\addlinespace
\textit{close receptacle} & close dishwasher 1 & You close dishwasher. \\
\addlinespace
\textit{turn on receptacle} & turn on dishwasher 1 & You turn on dishwasher. \\
\bottomrule
\end{tabularx}
\end{table*}

\begin{table*}[h!]
\caption{Representative action-observation pairs in AI2THOR under the partially observable setting.}
\label{tab:thor_obs_ex}
\centering
\small
\begin{tabularx}{\textwidth}{@{}p{2.3cm}p{2.6cm}X@{}}
\toprule
\textbf{Action Context} & \textbf{Action} & \textbf{Text Observation} \\
\midrule
\textit{initialization} & - & You are in the house, and you arrive at the bathroom, Looking quickly around the room, you see TowelHolder (1), Toilet (1), ToiletPaperHanger (1), Sink (1), SinkBasin (1), Shelf (1, 2, 3), HandTowelHolder (1), GarbageCan (1), BathtubBasin (1).\\
\addlinespace
\textit{go to object or receptacle} & go to BathtubBasin (1)  & You arrive at BathtubBasin (1). You see SoapBar (2), Cloth (1), Bathtub (1), BathtubBasin (1).  \\
\addlinespace
\textit{pick up object} & pick up SoapBar (2)  & You pick up SoapBar (2). You see SoapBar (2), Cloth (1), Bathtub (1), BathtubBasin (1). I am holding SoapBar (2) now. \\
\addlinespace
\textit{put down object} & put down SoapBar (2)  &  You put SoapBar (2) on Shelf (1). You see SoapBar (1, 2, 3), Shelf (1), Candle (2). \\
\addlinespace
\textit{slice object} & slice Potato (2) & You slice Potato (2). You see Potato (8, 9, 10, 11, 12), Knife (1), Fridge (1). Potato (5, 6, 7, 8, 9, 10, 11, 12) are sliced. I am holding Knife (1) now. \\
\addlinespace
\textit{open receptacle} & open Microwave (1)  & You open Microwave (1). You see StoveKnob (1, 2, 3, 4), StoveBurner (3), Potato (1), Pan (1, 2), Microwave (1), Bowl (3).\\
\addlinespace
\textit{close receptacle} & close Microwave (1)  &  You close Microwave (1). You see StoveKnob (1, 2, 3, 4), StoveBurner (3), Pan (1, 2), Microwave (1). \\
\addlinespace
\textit{turn on receptacle} &  turn on Microwave (1) & You turn on Microwave (1). You see Microwave (1), Drawer (5, 6), Cabinet (7, 8, 10). \\
\addlinespace
\textit{turn off receptacle} &  turn off Microwave (1) &  You turn off Microwave (1). You see Microwave (1), Drawer (5, 6), Cabinet (7, 8, 10). \\
\bottomrule
\end{tabularx}
\end{table*}
\clearpage

\section{Implementation Details of Baselines}
\label{app:baselines_details}
All baseline methods are evaluated under the same few-shot in-context learning setup as \emph{ReAcTree}, without fine-tuning. They also share identical retrieval size and decision cap: 5K tokens and 200 steps for WAH-NL, and 5K tokens and 100 steps for ALFRED. Details of episodic memory construction, generation strategies, and other implementation aspects are described for each baseline below.

\textbf{\emph{ZSP} [19].}
For WAH-NL, \emph{ZSP} constructs episodic memory using a fully observable, rule-based planner to extract 183 ground-truth trajectories from the training set. At test time, it retrieves in-context examples most similar to the current task goal using Sentence-BERT [34], following the shared 5K-token retrieval setup. The LLM performs free-form generation with temperature 0.0 to produce an acting action, which is then mapped to an executable skill via Sentence-BERT-based matching.  \emph{ZSP} generates its full plan recursively without access to intermediate observations, relying solely on the initial environment description. To compensate, we provide global environment information including all rooms, objects, and receptacles at the beginning of each task. Since object perception is unavailable at each step, we relax action constraints by omitting instance identifiers for \textit{pick up} and \textit{put down} actions, for example using \textit{pick up apple} instead of \textit{pick up apple 1}.

\textbf{\emph{Tree-Planner} [18].}
For WAH-NL, \emph{Tree-Planner} follows a two-step procedure consisting of plan sampling and grounded deciding. During plan sampling, it uses the same episodic memory as \emph{ZSP} (183 rule-based trajectories), retrieves up to 5K tokens of in-context examples, and generates $N$ candidate plans ($N=25$ or $50$) with temperature 0.8 and top-$p=0.95$, following the original paper. Similar to \emph{ZSP}, it also receives global environment information and generates plans without instance-level identifiers for \textit{pick up} and \textit{put down} actions, resulting in an action tree. In the grounded deciding phase, the model receives step-wise observations and selects an action from the pre-generated tree. This decision process uses a fixed set of in-context examples and leverages the Guidance library~[30] to perform constrained generation.

\textbf{\emph{ReAct} [55] \& \emph{ReAct+WM}.}
Both methods follow the same implementation settings as \emph{ReAcTree} in terms of episodic memory construction (manual collection followed by LLM-based bootstrapping), retrieval size and decision cap (5K and 200 for WAH-NL; 5K and 100 for ALFRED), and generation settings using the Guidance library~[30] with temperature 0.0 for reasoning and constrained action selection.

\section{Language Models}
\label{app:lang_models}

Table~\ref{tab:lang_models} lists the specific language models used in our experiments. Model names follow their identifiers in the HuggingFace model hub.

\begin{table}[h]
\caption{Language models used in our experiments.}
 \label{tab:lang_models}
 \centering
\begin{tabular}{@{}lll@{}}
\toprule
Class & Model name & Model size \\ \midrule 
\multirow{2}{*}{LLaMA 3.1}      & meta-llama/Llama-3.1-8B        & 8B   \\
                                & meta-llama/Llama-3.1-70B       & 70B  \\ \midrule
\multirow{2}{*}{Qwen 2.5}        & Qwen/Qwen2.5-7B               & 7B   \\
                                & Qwen/Qwen2.5-72B               & 72B  \\ \midrule
Mistral                         & mistralai/Mistral-7B-v0.3      & 7B   \\ \midrule
Gemma 2                           & google/gemma-2-9b              & 9B   \\ \midrule
Phi-4-reasonin-plus             & microsoft/Phi-4-reasoning-plus & 14B   \\
\bottomrule
\end{tabular}
\end{table}

\section{Prompts Templates}
\label{app:prompts}
\subsection{\emph{ReAcTree} \& \emph{ReAcTree+WM}}
\label{app:reactree_prompts}
We provide the prompt template of \emph{ReAcTree+WM} for WAH-NL and ALFRED in Listings~\ref{lst:prompt_template_reactree_wm_wah} and~\ref{lst:prompt_template_reactree_wm_alfred}, respectively. The only difference between \emph{ReAcTree+WM} and \emph{ReAcTree} lies in the omission of the ``\textit{}{recall location of}'' action, while otherwise sharing the same structure. The ``\textit{...}'' symbol denotes the position where in-context examples retrieved from episodic memory are inserted.

\begin{lstlisting}[caption={Prompt template of \emph{ReAcTree+WM} for WAH-NL}, label={lst:prompt_template_reactree_wm_wah}]
You are an advanced robot with ability to think, act, and expand behavior tree nodes in decision-making process. You can perform one of the following tasks:
1. Think: Use reasoning to satisfy the current goal condition.
2. Act: Execute a specific action to accomplish the current goal condition. You should use one of actions of this list: [go to, pick up, put down, open, close, turn on, recall location of, done, failure]
3. Expand: Decompose the current goal condition into more detailed subgoals. When expanding, generate appropriate control flow and subgoals. Control flow can be "sequence" (achieve subgoals sequentially; if any subgoal fails, the sequence is interrupted), "fallback" (attempt subgoals in order until one succeeds; if a subgoal is successful, the remaining subgoals are not attempted), or "parallel" (achieve subgoals in parallel; this enables tasks to continue independently, even if one subgoal fails).

Source domain:
...

Target_domain:
Your task is to: Put one cupcake and one apple on the coffee table
You are in the house, and there are 4 rooms: bathroom (1), bedroom (1), kitchen (1), living room (1). You are in the middle of a bedroom (1). Looking quickly around the room, you see bed (1), bookshelf (1), cabinet (1), chair (5), desk (1), nightstand (3).
\end{lstlisting}

\begin{lstlisting}[caption={Prompt template of \emph{ReAcTree+WM} for ALFRED}, label={lst:prompt_template_reactree_wm_alfred}]
You are an advanced robot with ability to think, act, and expand behavior tree nodes in decision-making process. You can perform one of the following tasks:
1. Think: Use reasoning to satisfy the current goal condition.
2. Act: Execute a specific action to accomplish the current goal condition. You should use one of actions of this list: [go to, pick up, put down, slice, open, close, turn on, turn off, done, failure]
3. Expand: Decompose the current goal condition into more detailed subgoals. When expanding, generate appropriate control flow and subgoals. Control flow can be "sequence" (achieve subgoals sequentially. If any subgoal fails, the sequence is interrupted) or "fallback" (Attempt subgoals in order until one succeeds. If a subgoal is successful, the remaining subgoals are not attempted).

Source domain:
...

Target domain:
Your task is to: Hold the clock and turn on the lamp.
You are in the house, and you arrive at the bedroom, Looking quickly around the room, you see Shelf (1, 2, 3, 4, 5, 6), Safe (1), LaundryHamper (1), GarbageCan (1), Drawer (1, 2, 3, 4, 5, 6), Desk (1, 2), Bed (1).
\end{lstlisting}

\subsection{Baselines}
\label{app:baselines_prompts}

We provide the prompt templates used for the baseline methods \emph{ZSP}, \emph{Tree-Planner}, and \emph{ReAct+WM} across both WAH-NL and ALFRED.

\textbf{\emph{ZSP}.} \emph{ZSP} uses the prompt template in Listing~\ref{lst:prompt_template_zsp}. The ``\textit{...}'' portion is replaced by in-context examples relevant to the current task goal.

\textbf{\emph{Tree-Planner}.} During its \emph{plan sampling} phase, \emph{Tree-Planner} employs the same template as \emph{ZSP} (Listing~\ref{lst:prompt_template_zsp}). For \emph{grounded deciding}, it uses a separate template in Listing~\ref{lst:prompt_template_treeplanner_gd}.

\textbf{\emph{ReAct} \& \emph{ReAct + WM}.} We provide the prompt templates of \emph{ReAct+WM} for WAH-NL and ALFRED in Listing~\ref{lst:prompt_template_react_wm_wah} and~\ref{lst:prompt_template_react_wm_alfred}, respectively. The only difference from \emph{ReAct} is the inclusion of the action ``\textit{recall location of}'' action from the prompt, while otherwise sharing the same structure. The ``\textit{...}'' portion in the template indicates where in-context examples retrieved from episodic memory are inserted.

\begin{lstlisting}[caption={Prompt template of \emph{ZSP} for WAH-NL}, label={lst:prompt_template_zsp}]
You are an advanced robot with ability to generate action plans. You can perform one of the following actions of this list: [go to, pick up, put down, open, close, turn on, done]

Source domain:
...

Target_domain:
You are in the house, and there are 4 rooms: bathroom (1), bedroom (1), kitchen (1), living room (1). In the kitchen 1, there are chair (1, 2, 3, 4), dishwasher (1), faucet (1), fridge (1), kitchen cabinet (1, 2, 3, 4, 5, 6, 7, 8), kitchen counter (1, 3), kitchen table (1), microwave oven (1), nightstand (1, 2), sink (1), sofa (1), stove (1), toaster (1). In the bathroom 1, there are bathroom cabinet (1), bathroom counter (1), faucet (2), sink (2), toilet (1), washing machine (1). In the bedroom 1, there are bed (1), bookshelf (1), cabinet (1), chair (5), desk (1), nightstand (3). In the living room 1, there are bookshelf (2, 3), chair (6), closet (1), coffee table (1), computer (1), desk (2), nightstand (4), sofa (2, 3), tv (1). The objects in the house are apple (1, 2, 3), bananas (1), bar soap (1), book (1, 2), bowl (1, 2, 3, 4, 5, 6, 7, 8, 9), box (1, 2, 3, 4, 5), bucket (1), candle (1), candy bar (1), cell phone (1, 2), cereal (1), chair (1, 2, 3, 4, 5, 6), chips (1, 2), chocolate syrup (1), coffee pot (1), condiment bottle (1, 2, 3, 4), condiment shaker (1, 2, 3, 4), cooking pot (1, 2), crackers (1, 2), creamy buns (1), cupcake (1, 2, 3), cutlery fork (1, 2, 3, 4), cutlery knife (1, 2, 3, 4, 5, 6), cutlets (1), dishwashing liquid (1), face cream (1, 2, 3), folder (1, 2, 3, 4), frying pan (1), hair product (1, 2, 3, 4), hanger (1, 2, 3, 4, 5, 6, 7), juice (1, 2), keyboard (1), lime (1), milk (1), mouse (1), mug (1, 2, 3), notes (1), oven tray (1), pancake (1), pear (1), pile of clothes (1, 2), pillow (1, 2, 3, 4, 5, 6), plate (1, 2, 3, 4, 5, 6, 7), plum (1, 2), pudding (1, 2), radio (1), rug (1, 2, 3, 4), slice of bread (1, 2), toilet paper (1), toothbrush (1), toothpaste (1), wall phone (1), wall picture frame (1, 2, 3, 4, 5, 6, 7, 8), washing sponge (1), water glass (1, 2, 3, 4, 5), wine (1). You are in the middle of a bedroom (1). Looking quickly around the room, you see bed (1), bookshelf (1), cabinet (1), chair (5), desk (1), nightstand (3).
Your task is to: Put one cupcake and one apple on the coffee table
\end{lstlisting}

\begin{lstlisting}[caption={Prompt template of \emph{Tree-Planner} grounded deciding for WAH-NL}, label={lst:prompt_template_treeplanner_gd}]
You need to act as a home robot. At each moment, I will provide you with observations of your current environment, as well as the high-level task I want you to do, and previous mid-level sub-tasks that have been executed. Then, you need to select the best sub-task from the options I provide to complete the designated home task based on the observation and your past experience. When one choosed sub-task causes an error in the environment, you will be provided with the error information and the corresponding sub-task, and you need to re-choose a corrective sub-task at the current time step. For example, the actions that have been executed in the environment are:
go to kitchen 1
go to kitchen table 1
The choosed action is: pick up cutlery fork 1
The prompt (error information) would be: The action: "pick up cutlery fork 1" caused an error: Action is not executable, since the agent is not close to cutlery fork 1 when executing "pick up cutlery fork 1" Among the following actions, wich action would you take.
go to dishwasher 1
go to fridge 1
A corrective choice of action would be: go to dishwasher 1

Currently, you are in the house, and there are 4 rooms: bathroom (1), bedroom (1), kitchen (1), living room (1). You are in the middle of a bedroom (1). Looking quickly around the room, you see bed (1), bookshelf (1), cabinet (1), chair (5), desk (1), nightstand (3). You are close to cabinet (1), cell phone (2), desk (1), plate (7). You hold nothing in your hands.
Your task is to: Put one cupcake and one apple on the coffee table
\end{lstlisting}

\begin{lstlisting}[caption={Prompt template of \emph{ReAct+WM} for WAH-NL}, label={lst:prompt_template_react_wm_wah}]
You are an advanced robot with ability to think and act. You can perform one of the following tasks:
1. Think: Use reasoning to satisfy the current goal condition.
2. Act: Execute a specific action to accomplish the current goal condition. You should use one of actions of this list: [go to, pick up, put down, open, close, turn on, recall location of, done, failure]
Source domain:
...

Target_domain:
Your task is to: Put one cupcake and one apple on the coffee table
You are in the house, and there are 4 rooms: bathroom (1), bedroom (1), kitchen (1), living room (1). You are in the middle of a bedroom (1). Looking quickly around the room, you see bed (1), bookshelf (1), cabinet (1), chair (5), desk (1), nightstand (3).
\end{lstlisting}

\begin{lstlisting}[caption={Prompt template of ReAct+WM for ALFRED}, label={lst:prompt_template_react_wm_alfred}]
You are an advanced robot with ability to think and act. You can perform one of the following tasks:
1. Think: Use reasoning to satisfy the current goal condition.
2. Act: Execute a specific action to accomplish the current goal condition. You should use one of actions of this list: [go to, pick up, put down, slice, open, close, turn on, turn off, done, failure]

Source domain:
...

Target domain:
Your task is to: Hold the clock and turn on the lamp.
You are in the house, and you arrive at the bedroom, Looking quickly around the room, you see Shelf (1, 2, 3, 4, 5, 6), Safe (1), LaundryHamper (1), GarbageCan (1), Drawer (1, 2, 3, 4, 5, 6), Desk (1, 2), Bed (1).
\end{lstlisting}

\section{Failure Case of \emph{ReAct+WM}}
\label{app:react_fail_ex}

\begin{figure*}[h!]
\centering
\includegraphics[width=0.9\textwidth]{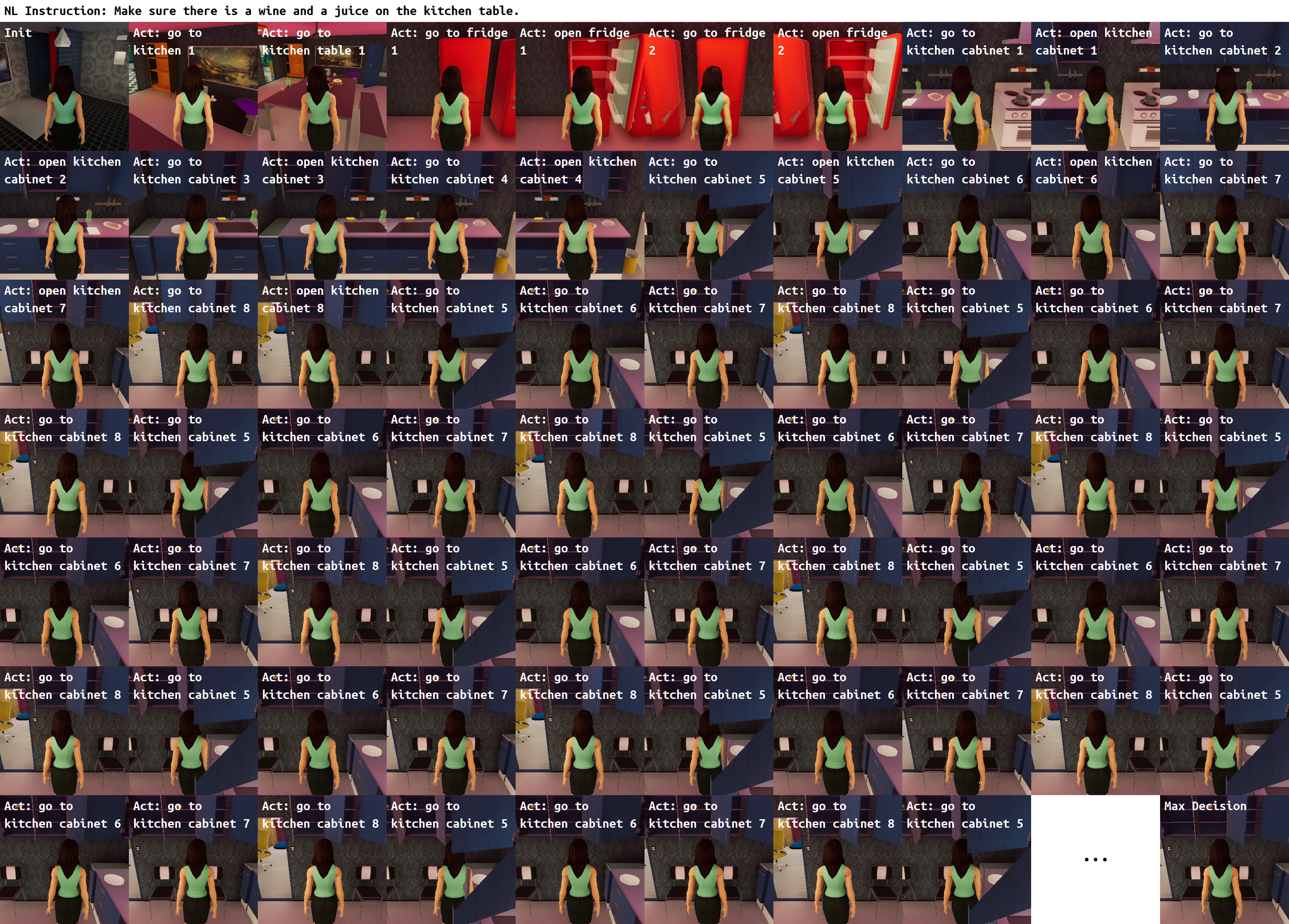}
\caption{Failure case of \emph{ReAct+WM} on the WAH-NL dataset using Qwen~2.5~72B.}
\label{fig:react_failure_ex}
\end{figure*}

\clearpage

\section{\emph{ReAcTree+WM} Full Trajectory of Success Case}
\label{app:success_traj_reactree_wm}
\begin{lstlisting}[caption={Full Trajectory of ReAcTree+WM}, label={lst:prompt_react}]

### Agent Node 0
Your task is to: Make sure there is a wine and a juice on the coffee table.
You are in the house, and there are 4 rooms: bathroom (1), bedroom (1), kitchen (1), living room (1). You are in the middle of a bathroom (1). Looking quickly around the room, you see bathroom cabinet (1), bathroom counter (1), faucet (1), sink (1), toilet (1), towel rack (1), washing machine (1).
Think: To solve the task, I need to accomplish 2 subgoals independently. Move the wine on the coffee table, and move the juice on the coffee table.
Expand: {'control_flow': 'parallel', 'conditions': 'move the wine on the coffee table, and move the juice on the coffee table'}

### Agent 1
Your primary goal is to: Make sure there is a wine and a juice on the coffee table.
To achieve this, you should perform your sibling tasks in parallel. At this level, your sibling tasks are: move the wine on the coffee table, and and move the juice on the coffee table.
Your task is to: move the wine on the coffee table
You are in the house, and there are 4 rooms: bathroom (1), bedroom (1), kitchen (1), living room (1). You are in the middle of a bathroom (1). Looking quickly around the room, you see bathroom cabinet (1), bathroom counter (1), faucet (1), sink (1), toilet (1), towel rack (1), washing machine (1).
Think: To solve the task, I need to accomplish 2 subgoals sequentially. Find and pick up the wine, and find the coffee table and put down the wine.
Expand: {'control_flow': 'sequence', 'conditions': 'find and pick up the wine, find the coffee table and put down the wine'}

### Agent Node 3
Your primary goal is to: move the wine on the coffee table
To achieve this, you should perform your sibling tasks in sequence. At this level, your sibling tasks are: find and pick up the wine, and find the coffee table and put down the wine.
Your task is to: find and pick up the wine
You are in the house, and there are 4 rooms: bathroom (1), bedroom (1), kitchen (1), living room (1). You are in the middle of a bathroom (1). Looking quickly around the room, you see bathroom cabinet (1), bathroom counter (1), faucet (1), sink (1), toilet (1), towel rack (1), washing machine (1).
Think: To solve the task, I need to recall the location of the wine.
Act: recall location of wine
You have not seen wine before.
Think: Now, I need to find and pick up the wine in one of the rooms. The wine is more likely to appear in kitchen 1, living room 1, or bedroom 1. I can check each room one by one until I find and pick up the wine.
Expand: {'control_flow': 'fallback', 'conditions': 'find and pick up the wine in kitchen 1, find and pick up the wine in living room 1, find and pick up the wine in bedroom 1'}

### Agent Node 5
Your primary goal is to: find and pick up the wine
To achieve this, you should perform your sibling tasks using a fallback strategy. At this level, your sibling tasks are: find and pick up the wine in kitchen 1, find and pick up the wine in living room 1, and find and pick up the wine in bedroom 1.
Your task is to: find and pick up the wine in kitchen 1
You are in the house, and there are 4 rooms: bathroom (1), bedroom (1), kitchen (1), living room (1). You are in the middle of a bathroom (1). Looking quickly around the room, you see bathroom cabinet (1), bathroom counter (1), faucet (1), sink (1), toilet (1), towel rack (1), washing machine (1).
Think: I am in bathroom 1. Now, I need to go to kitchen 1.
Act: go to kitchen 1
You move to the kitchen (1). Looking quickly around the room, you see bench (1, 2), bookshelf (1, 2), chair (1, 2), dishwasher (1), faucet (2), fridge (1, 2), garbage can (1), kitchen cabinet (1, 2, 3, 4, 5, 6, 7, 8), kitchen counter (1, 2), kitchen table (1), microwave oven (1), sink (2), stove (1), toaster (1).
Think: Now, I need to find the wine. The wine is more likely to appear on kitchen table (1), or in fridge (1, 2), kitchen cabinet (1, 2, 3, 4, 5, 6, 7, 8). I can check each place one by one, starting from kitchen table 1, until I find the wine.
Act: go to kitchen table 1
You arrive at the kitchen table (1). You see bench (2), book (1), bowl (6), cereal (1), kitchen table (1), milk (1), pudding (1, 2)
Act: go to fridge 1
You arrive at the fridge (1). The fridge (1) is closed. You see bench (1), cupcake (1), fridge (1)
Act: open fridge 1
You open fridge. You see bench (1), cupcake (1), fridge (1)
Act: go to fridge 2
You arrive at the fridge (2). The fridge (2) is closed. You see bench (1), cupcake (1), fridge (1, 2), juice (1)
Act: open fridge 2
You open fridge. You see bench (1), cupcake (1), fridge (1, 2), juice (1)
Act: go to kitchen cabinet 1
You arrive at the kitchen cabinet (1). The kitchen cabinet (1) is closed. You see chicken (1), cooking pot (1), cutlery knife (1, 2), cutlets (1), dishwashing liquid (1), faucet (2), frying pan (1), garbage can (1), kitchen cabinet (1, 4), kitchen counter (1), sink (2), stove (1)
Act: open kitchen cabinet 1
You open kitchen cabinet. You see chicken (1), cooking pot (1), cutlery knife (1, 2), cutlets (1), dishwashing liquid (1), faucet (2), frying pan (1), garbage can (1), kitchen cabinet (1, 4), kitchen counter (1), sink (2), stove (1)
Act: go to kitchen cabinet 2
You arrive at the kitchen cabinet (2). The kitchen cabinet (2) is closed. You see bowl (1, 2, 3, 4, 5), chair (2), chicken (1), cutlery knife (1, 2), cutlets (1), dishwashing liquid (1), faucet (2), kitchen cabinet (1, 2, 3, 4, 5, 6), kitchen counter (1), plate (1, 2, 3, 4), sink (2), washing sponge (1)
Act: open kitchen cabinet 2
You open kitchen cabinet. You see bowl (1, 2, 3, 4, 5), chair (2), chicken (1), cutlery knife (1, 2), cutlets (1), dishwashing liquid (1), faucet (2), kitchen cabinet (1, 2, 3, 4, 5, 6), kitchen counter (1), plate (1, 2, 3, 4), sink (2), washing sponge (1)
Act: go to kitchen cabinet 3
You arrive at the kitchen cabinet (3). The kitchen cabinet (3) is closed. You see bowl (1, 2, 3, 4, 5), chair (1, 2), dishwashing liquid (1), faucet (2), kitchen cabinet (2, 3, 4, 5, 6, 7), kitchen counter (1, 2), plate (1, 2, 3, 4), sink (2), washing sponge (1)
Act: open kitchen cabinet 3
You open kitchen cabinet. You see bowl (1, 2, 3, 4, 5), chair (1, 2), dishwashing liquid (1), faucet (2), kitchen cabinet (2, 3, 4, 5, 6, 7), kitchen counter (1, 2), plate (1, 2, 3, 4), sink (2), washing sponge (1)
Act: go to kitchen cabinet 4
You arrive at the kitchen cabinet (4). The kitchen cabinet (4) is closed. You see bowl (1, 2, 3, 4, 5), chicken (1), cooking pot (1), cutlery knife (1, 2), cutlets (1), dishwashing liquid (1), faucet (2), garbage can (1), kitchen cabinet (1, 2, 4), kitchen counter (1), sink (2), washing sponge (1)
Act: open kitchen cabinet 4
You open kitchen cabinet. You see bowl (1, 2, 3, 4, 5), chicken (1), cooking pot (1), cutlery knife (1, 2), cutlets (1), dishwashing liquid (1), faucet (2), garbage can (1), kitchen cabinet (1, 2, 4), kitchen counter (1), sink (2), washing sponge (1)
Act: go to kitchen cabinet 5
You arrive at the kitchen cabinet (5). The kitchen cabinet (5) is closed. You see bowl (1, 2, 3, 4, 5), chair (1, 2), cutlets (1), dishwashing liquid (1), faucet (2), kitchen cabinet (2, 3, 4, 5, 6, 7), kitchen counter (1, 2), plate (1, 2, 3, 4), sink (2), washing sponge (1)
Act: open kitchen cabinet 5
You open kitchen cabinet. You see bowl (1, 2, 3, 4, 5), chair (1, 2), cutlets (1), dishwashing liquid (1), faucet (2), kitchen cabinet (2, 3, 4, 5, 6, 7), kitchen counter (1, 2), plate (1, 2, 3, 4), sink (2), washing sponge (1)
Act: go to kitchen cabinet 6
You arrive at the kitchen cabinet (6). The kitchen cabinet (6) is closed. You see bowl (1, 2, 3, 4, 5), chair (1, 2), dishwashing liquid (1), faucet (2), kitchen cabinet (2, 3, 4, 5, 6, 7, 8), kitchen counter (1, 2), plate (1, 2, 3, 4), sink (2), washing sponge (1)
Act: open kitchen cabinet 6
You open kitchen cabinet. You see bowl (1, 2, 3, 4, 5), chair (1, 2), dishwashing liquid (1), faucet (2), kitchen cabinet (2, 3, 4, 5, 6, 7, 8), kitchen counter (1, 2), plate (1, 2, 3, 4), sink (2), washing sponge (1)
Act: go to kitchen cabinet 7
You arrive at the kitchen cabinet (7). The kitchen cabinet (7) is closed. You see bowl (1, 2, 3, 4, 5), chair (1, 2), kitchen cabinet (2, 3, 4, 5, 6, 7, 8), kitchen counter (1), light switch (2), plate (1, 2, 3, 4), sink (2)
Act: open kitchen cabinet 7
You open kitchen cabinet. You see bowl (1, 2, 3, 4, 5), chair (1, 2), kitchen cabinet (2, 3, 4, 5, 6, 7, 8), kitchen counter (1), light switch (2), plate (1, 2, 3, 4), pudding (3), sink (2)
Act: go to kitchen cabinet 8
You arrive at the kitchen cabinet (8). The kitchen cabinet (8) is closed. You see bench (2), chair (1, 2), kitchen cabinet (5, 6, 7, 8), light switch (2), pudding (3)
Act: open kitchen cabinet 8
You open kitchen cabinet. You see bench (2), chair (1, 2), kitchen cabinet (5, 6, 7, 8), light switch (2), pudding (3)
Think: I cannot find the wine in kitchen 1.
Act: failure

### Agent Node 6
Your primary goal is to: find and pick up the wine
To achieve this, you should perform your sibling tasks using a fallback strategy. At this level, your sibling tasks are: find and pick up the wine in kitchen 1, find and pick up the wine in living room 1, and find and pick up the wine in bedroom 1.
Your task is to: find and pick up the wine in living room 1
You are in the house, and there are 4 rooms: bathroom (1), bedroom (1), kitchen (1), living room (1). You are in the middle of a kitchen (1). Looking quickly around the room, you see bench (1, 2), bookshelf (1, 2), chair (1, 2), dishwasher (1), faucet (2), fridge (1, 2), garbage can (1), kitchen cabinet (1, 2, 3, 4, 5, 6, 7, 8), kitchen counter (1, 2), kitchen table (1), microwave oven (1), sink (2), stove (1), toaster (1).
Think: I am in kitchen 1. Now, I need to go to living room 1.
Act: go to living room 1
You move to the living room (1). Looking quickly around the room, you see bookshelf (3), chair (3), closet (1), coffee table (1), computer (1), desk (1), garbage can (2), nightstand (1), sofa (1), tv (1).
Think: Now, I need to find the wine. The wine is more likely to appear on coffee table (1), desk (1), or in closet (1), nightstand (1). I can check each place one by one, starting from coffee table 1, until I find the wine.
Act: go to coffee table 1
You arrive at the coffee table (1). You see alcohol (2), bananas (1), bowl (7), cell phone (1, 2), coffee table (1), lime (1), mug (1), plum (1, 2), sofa (1)
Act: go to desk 1
You arrive at the desk (1). You see chair (3), computer (1), desk (1), nightstand (1), water glass (2)
Act: go to closet 1
You arrive at the closet (1). The closet (1) is closed. You see closet (1), light switch (3)
Act: open closet 1
You open closet. You see closet (1), light switch (3)
Act: go to nightstand 1
You arrive at the nightstand (1). The nightstand (1) is closed. You see box (2, 5, 7), chair (3), computer (1), desk (1), nightstand (1), water glass (2)
Act: open nightstand 1
You open nightstand. You see box (2, 5, 7), chair (3), computer (1), desk (1), nightstand (1), water glass (2)
Think: I cannot find the wine in living room 1.
Act: failure

### Agent Node 7
Your primary goal is to: find and pick up the wine
To achieve this, you should perform your sibling tasks using a fallback strategy. At this level, your sibling tasks are: find and pick up the wine in kitchen 1, find and pick up the wine in living room 1, and find and pick up the wine in bedroom 1.
Your task is to: find and pick up the wine in bedroom 1
You are in the house, and there are 4 rooms: bathroom (1), bedroom (1), kitchen (1), living room (1). You are in the middle of a living room (1). Looking quickly around the room, you see bookshelf (3), chair (3), closet (1), coffee table (1), computer (1), desk (1), garbage can (2), nightstand (1), sofa (1), tv (1).
Think: I am in living room 1. Now, I need to go to bedroom 1.
Act: go to bedroom 1
You move to the bedroom (1). Looking quickly around the room, you see bed (1), bookshelf (4), cabinet (1), chair (4), closet (2), coffee table (2), computer (2), desk (2, 3), garbage can (3).
Think: Now, I need to find the wine. The wine is more likely to appear on coffee table (2), or in cabinet (1), closet (2). I can check each place one by one, starting from coffee table 2, until I find the wine.
Act: go to coffee table 2
You arrive at the coffee table (2). You see coffee table (2), magazine (1)
Act: go to cabinet 1
You arrive at the cabinet (1). The cabinet (1) is closed. You see cabinet (1), computer (2), desk (2), light switch (4), sundae (1)
Act: open cabinet 1
You open cabinet. You see cabinet (1), computer (2), desk (2), light switch (4), sundae (1), wine (1)
Think: I find wine 1. Now I need to pick it up.
Act: pick up wine 1
You pick up wine. You hold wine (1).
Think: Now I accomplish the goal.
Act: done

### Agent Node 4
Your primary goal is to: move the wine on the coffee table
To achieve this, you should perform your sibling tasks in sequence. At this level, your sibling tasks are: find and pick up the wine, and find the coffee table and put down the wine.
Your task is to: find the coffee table and put down the wine
You are in the house, and there are 4 rooms: bathroom (1), bedroom (1), kitchen (1), living room (1). You are in the middle of a bedroom (1). Looking quickly around the room, you see bed (1), bookshelf (4), cabinet (1), chair (4), closet (2), coffee table (2), computer (2), desk (2, 3), garbage can (3). You hold wine (1).
Think: I hold wine 1. Now, I need to find the coffee table. The coffee table is more likely to appear in living room 1.
Act: go to living room 1
You move to the living room (1). Looking quickly around the room, you see bookshelf (3), chair (3), closet (1), coffee table (1), computer (1), desk (1), garbage can (2), nightstand (1), sofa (1), tv (1). You hold wine (1).
Think: I can see coffee table 1.
Act: go to coffee table 1
You arrive at the coffee table (1). You see alcohol (2), bananas (1), bowl (7), cell phone (1, 2), coffee table (1), lime (1), mug (1), plum (1, 2), sofa (1), wine (1) You hold wine (1).
Think: Now, I need to put down wine 1.
Act: put down wine 1
You put down wine on coffee table
Think: Now I accomplish the goal.
Act: done

### Agent Node 2
Your primary goal is to: Make sure there is a wine and a juice on the coffee table.
To achieve this, you should perform your sibling tasks in parallel. At this level, your sibling tasks are: move the wine on the coffee table, and and move the juice on the coffee table.
Your task is to: and move the juice on the coffee table
You are in the house, and there are 4 rooms: bathroom (1), bedroom (1), kitchen (1), living room (1). You are in the middle of a living room (1). Looking quickly around the room, you see bookshelf (3), chair (3), closet (1), coffee table (1), computer (1), desk (1), garbage can (2), nightstand (1), sofa (1), tv (1).
Think: To solve the task, I need to accomplish 2 subgoals sequentially. Find and pick up the juice, and find the coffee table and put down the juice.
Expand: {'control_flow': 'sequence', 'conditions': 'find and pick up the juice, find the coffee table and put down the juice'}

### Agent Node 8
Your primary goal is to: and move the juice on the coffee table
To achieve this, you should perform your sibling tasks in sequence. At this level, your sibling tasks are: find and pick up the juice, and find the coffee table and put down the juice.
Your task is to: find and pick up the juice
You are in the house, and there are 4 rooms: bathroom (1), bedroom (1), kitchen (1), living room (1). You are in the middle of a living room (1). Looking quickly around the room, you see bookshelf (3), chair (3), closet (1), coffee table (1), computer (1), desk (1), garbage can (2), nightstand (1), sofa (1), tv (1).
Think: To solve the task, I need to recall the location of the juice
Act: recall location of juice
You saw juice 1 near fridge 2 in kitchen 1.
Think: I can find juice 1 near fridge 2 in kitchen 1. I can see fridge 2.
Act: go to kitchen 1
You move to the kitchen (1). Looking quickly around the room, you see bench (1, 2), bookshelf (1, 2), chair (1, 2), dishwasher (1), faucet (2), fridge (1, 2), garbage can (1), kitchen cabinet (1, 2, 3, 4, 5, 6, 7, 8), kitchen counter (1, 2), kitchen table (1), microwave oven (1), sink (2), stove (1), toaster (1).
Think: I can see fridge 2.
Act: go to fridge 2
You arrive at the fridge (2). The fridge (2) is open. You see bench (1), cupcake (1), fridge (1, 2), juice (1)
Think: I find juice 1. Now I need to pick it up.
Act: pick up juice 1
You pick up juice. You hold juice (1).
Think: Now I accomplish the goal.
Act: done

### Agent Node 9
Your primary goal is to: and move the juice on the coffee table
To achieve this, you should perform your sibling tasks in sequence. At this level, your sibling tasks are: find and pick up the juice, and find the coffee table and put down the juice.
Your task is to: find the coffee table and put down the juice
You are in the house, and there are 4 rooms: bathroom (1), bedroom (1), kitchen (1), living room (1). You are in the middle of a kitchen (1). Looking quickly around the room, you see bench (1, 2), bookshelf (1, 2), chair (1, 2), dishwasher (1), faucet (2), fridge (1, 2), garbage can (1), kitchen cabinet (1, 2, 3, 4, 5, 6, 7, 8), kitchen counter (1, 2), kitchen table (1), microwave oven (1), sink (2), stove (1), toaster (1). You hold juice (1).
Think: I hold juice 1. Now, I need to find the coffee table. The coffee table is more likely to appear in living room 1.
Act: go to living room 1
You move to the living room (1). Looking quickly around the room, you see bookshelf (3), chair (3), closet (1), coffee table (1), computer (1), desk (1), garbage can (2), nightstand (1), sofa (1), tv (1). You hold juice (1).
Think: I can see coffee table 1.
Act: go to coffee table 1
You arrive at the coffee table (1). You see alcohol (2), bananas (1), bowl (7), cell phone (1, 2), coffee table (1), juice (1), lime (1), mug (1), plum (1, 2), sofa (1), wine (1) You hold juice (1).
Think: Now, I need to put down juice 1.
Act: put down juice 1
You put down juice on coffee table
Think: Now I accomplish the goal.
Act: done
\end{lstlisting}
\clearpage

\section{Qualitative Results for ALFRED}
\label{app:qualitative_results_for_alfred}
\subsection{Success Case of \emph{ReAcTree+WM} on ALFRED}

\begin{figure*}[h!]
\centering
\begin{minipage}{\textwidth}
\centering
\includegraphics[width=0.8\textwidth]{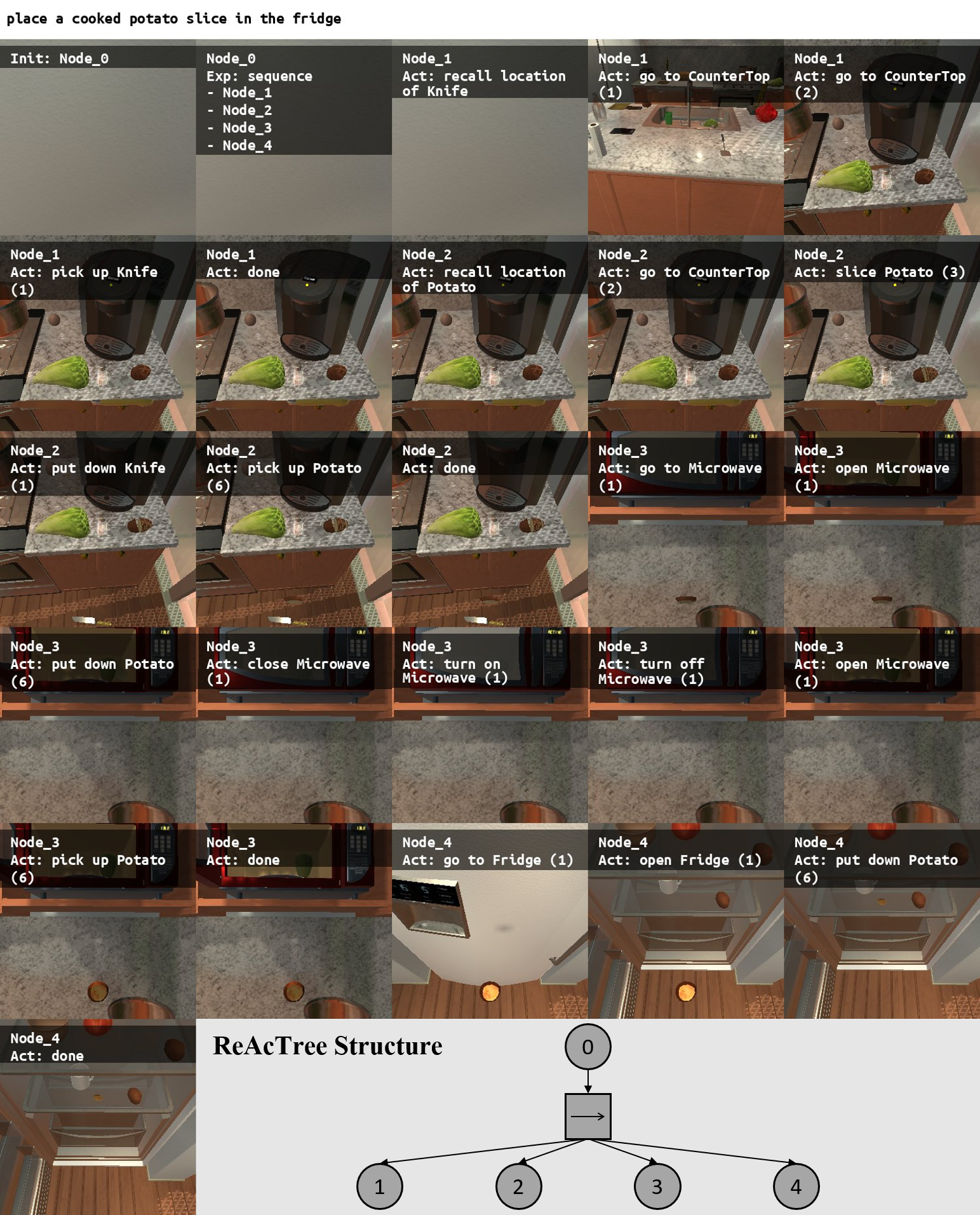}
\end{minipage}
\begin{minipage}{\textwidth}
    \centering
    {\begin{adjustbox}{width=0.8\textwidth}
    \begin{tabular}{ll||ll}
        Node 0 & \textit{Place a cooked potato slice in the fridge} & Node 3 & \textit{cook and pick up potato} \\
        Node 1 & \textit{find and pick up knife} & Node 4 & \textit{place potato in fridge}\\
        Node 2 & \textit{slice and pick up potato} &  & \\
    \end{tabular}
    \end{adjustbox}}
\end{minipage}
\caption{Success case of \emph{ReAcTree+WM} on ALFRED using LLaMA~3.1~70B.}
\label{fig:alfredd_reactree_success_ex}
\end{figure*}

\clearpage

\subsection{Failure Case of \emph{ReAct+WM} on ALFRED}

\begin{figure*}[h!]
\centering
\includegraphics[width=0.8\textwidth]{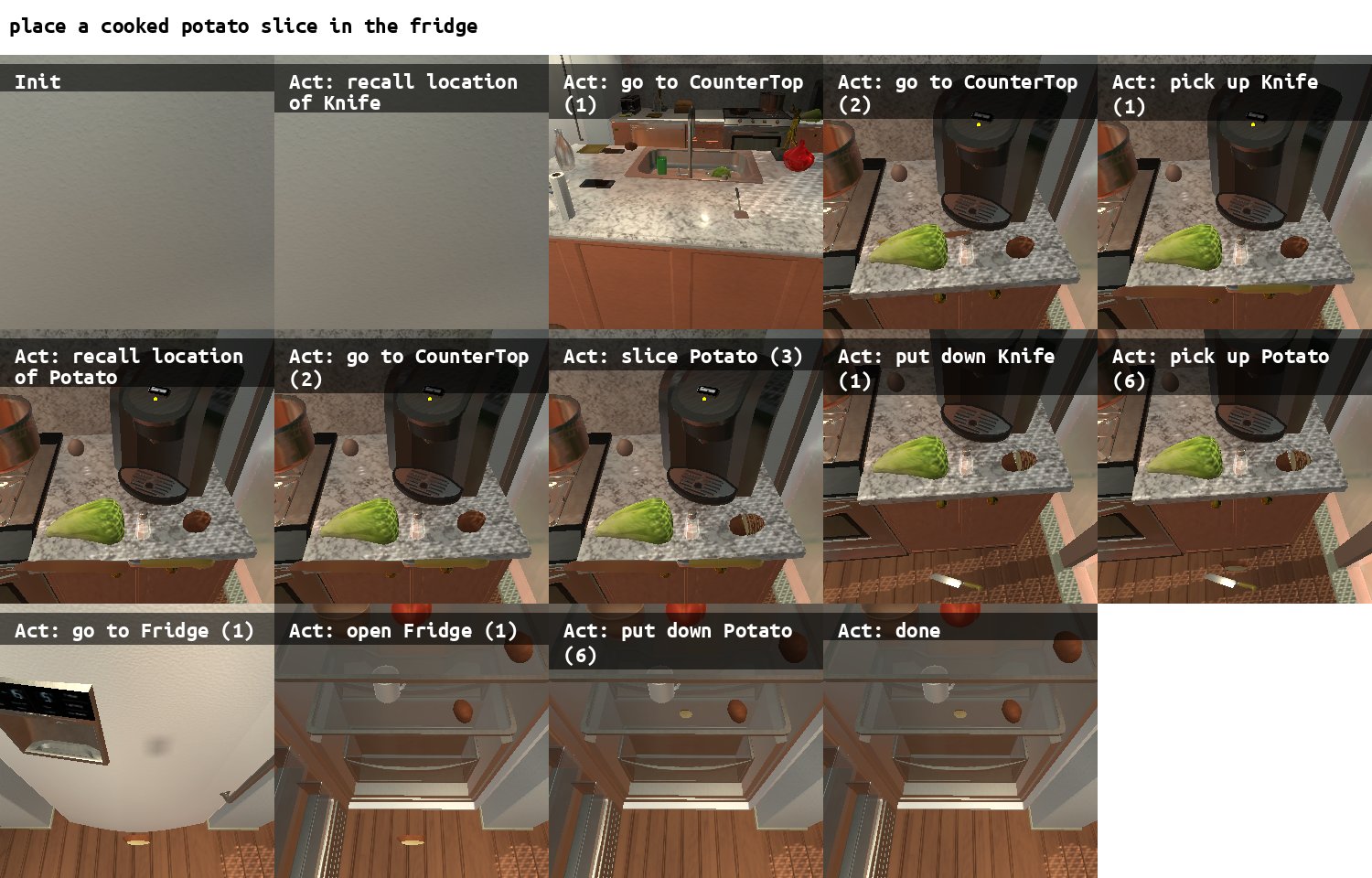}
\caption{Failure case of \emph{ReAct+WM} on ALFRED using LLaMA~3.1~70B.}
\label{fig:alfred_react_failure_ex}
\end{figure*}